\documentclass[notitlepage,letterpaper,12pt]{jedm}

\usepackage{amsmath}
\usepackage{amssymb}
\usepackage{booktabs}
\usepackage{graphicx}
\usepackage{array}
\usepackage{ragged2e}
\usepackage{rotating}
\usepackage{placeins}
\usepackage{url}
\usepackage[hidelinks]{hyperref}
\usepackage{tikz}
\usetikzlibrary{positioning,arrows.meta,calc}

\let\citep\cite
\let\citet\citeN

\newcommand{\dM}{d_{M}}
\newcommand{\dS}{d_{S}}
\newcommand{\sg}{\operatorname{sg}}

\graphicspath{{figures/}{./}}
\title{Prediction Is Not Calibration: Item Parameter Recovery and Representation Coupling in Knowledge Tracing}
\author{{\large Wenrui Yuan}\\
University of Twente\\Enschede, The Netherlands
\and
{\large St\'{e}phanie M. van den Berg}\\
University of Twente\\Enschede, The Netherlands
\and
{\large Johannes Steinr\"{u}cke}\\
University of Twente\\Enschede, The Netherlands}
\date{}

\begin{document}
\maketitle

\begin{abstract}
Knowledge tracing (KT) models can estimate item parameters through an item
response decoder while training solely for next-response prediction, yet high
predictive accuracy does not guarantee that the learned parameters recover the
underlying item parameters. We investigate item parameter estimation across 2PL, GPCM,
and NRM decoders paired with LSTM, Transformer, and DKVMN sequence encoders.
First, response-model derivatives show why discrimination and option slope
updates depend directly on the scale and spread of estimated ability, whereas
difficulty, step threshold, and intercept updates lack an ability multiplier outside the
prediction residual. Second, affine logit identities show how item-dependent shifts and
rescaling in estimated ability are absorbed into the estimated item parameters
without altering response probabilities. In controlled simulations, difficulty,
mean threshold, and correct-option intercept recovery remain high under shuffled order across item
representation widths, whereas discrimination and keyed option slope contrast
recovery require wider item representations and degrade further under fixed item
order. Separating the sequence item representation (width \(\dS\)) from the item
parameter key (width \(\dM\)) isolates these two pathways, and pairing a wide
parameter key with a thin sequence representation (\(\mathrm{SK}(64,8)\)) reduces
fixed-order ability distortions and improves parameter recovery and cross-order
consistency without sacrificing test accuracy or increasing model size relative
to a shared width-64 baseline (\(\mathrm{SH}(64)\)). On TIMSS and EdNet,
widening \(\dM\) consistently improves agreement with an offline IRT reference,
while the trade-off from narrowing \(\dS\) depends on how strongly the encoder
relies on sequence width for prediction. Together, these findings show why KT
item parameters, especially discrimination and option slopes, must be evaluated
separately from response prediction accuracy.
\\
{\parindent0pt
\textbf{Keywords:} knowledge tracing, item response theory, parameter estimation,
item representations, sequence models
}
\end{abstract}

\section{Introduction}\label{sec:intro}

Embedding an item response theory (IRT) equation inside a neural knowledge
tracing (KT) model is an appealing way to combine sequential prediction with
psychometric interpretability. Whereas a generic neural decoder maps a hidden
state to response probabilities through an unconstrained linear projection
\citep{piech_deep_2015,zhang_dkvmn_2017,pandey_sakt_2019,bai_survey_2024}, an
IRT decoder predicts responses through an explicit measurement equation that
pairs a dynamic learner ability estimate with item parameters. Examples
include the two-parameter logistic model (2PL) with discrimination and
difficulty, the generalized partial credit model (GPCM) with discrimination and
step thresholds, and the nominal response model (NRM) with option slopes and
intercepts
\citep{lord_1980,muraki_gpcm_1992,thissen_steinberg_1986,deayala_theory_2009}.
Because the sequence encoder, ability readout, and item parameter readout can be
trained end to end under next-response cross-entropy loss, the learned decoder
weights are often treated directly as calibrated item parameter estimates in sequential
prediction \citep{yeung_deepirt_2019,tsutsumi_independent_2021,vie_kashima_2023}.

Yet adopting the functional form of an IRT model at the output layer does not
guarantee that the learned weights recover the psychometric properties of the
items. In marginal maximum-likelihood IRT estimation, item parameters are identified against a
common latent scale because learner ability is treated as a random effect with a
specified prior across an entire response matrix \citep{bock_aitkin_1981}. In
sequential KT, by contrast, the scalar state readout \(\hat\theta^-_{i,t}\)
supplied to the decoder is constructed dynamically from the preceding
interaction history \(\mathcal H_{i,t-1}\) to minimize next-response loss. Prior
KT--IRT studies have largely left this identification problem open, either by
using one-parameter logistic (1PL) decoders with fixed discrimination and
evaluating only item difficulty
\citep{yeung_deepirt_2019,tsutsumi_independent_2021}, or by establishing an
algebraic correspondence between a KT decoder and an IRT model while evaluating
only predictive accuracy \citep{vie_kashima_2023}. Whether a sequential KT model
trained solely for next-response prediction can recover item discrimination,
polytomous step thresholds, or nominal option slopes, and under what
architectural conditions those estimates degrade, remains an open question.

In this paper, we show that next-response prediction and item parameter
recovery diverge for two structural reasons in the IRT
decoder and the sequence encoder.
\begin{enumerate}
\item \textbf{Asymmetry between location and slope updates.} In the first-order
derivatives of the 2PL, GPCM, and NRM prediction losses, updates to location
parameters (difficulty, step thresholds, and option intercepts) depend directly
on the prediction residual, whereas updates to slope parameters (discrimination
and option slopes) multiply the residual by the ability estimate or the
ability-to-location distance. Location parameters can therefore be anchored by
marginal response rates even when ability estimates are compressed or noisy,
whereas discrimination and option slopes are directly sensitive to the spread
and resolution of the item and ability representations.
\item \textbf{Affine invariance to item-specific ability shifts.} Because
ability and item parameters enter the response probability jointly, any
item-dependent shift \(\phi_q\) or positive rescaling \(s_q>0\) in the encoder's ability
estimates \(\hat\theta^-_{i,t}\) can be absorbed into the item parameters along
an affine null direction of the decoder logits without altering predicted
response probabilities. When items appear in a structured or fixed order, the
preceding history carries information about the upcoming item, allowing a
wide sequence representation to act as a predictive shortcut that encodes item-specific offsets and gains
into \(\hat\theta^-_{i,t}\). Moreover, when a single shared item embedding
table feeds both the sequence encoder and the item parameter readout, backward gradients
from later responses update the same item representation used to read out item
parameters.
\end{enumerate}

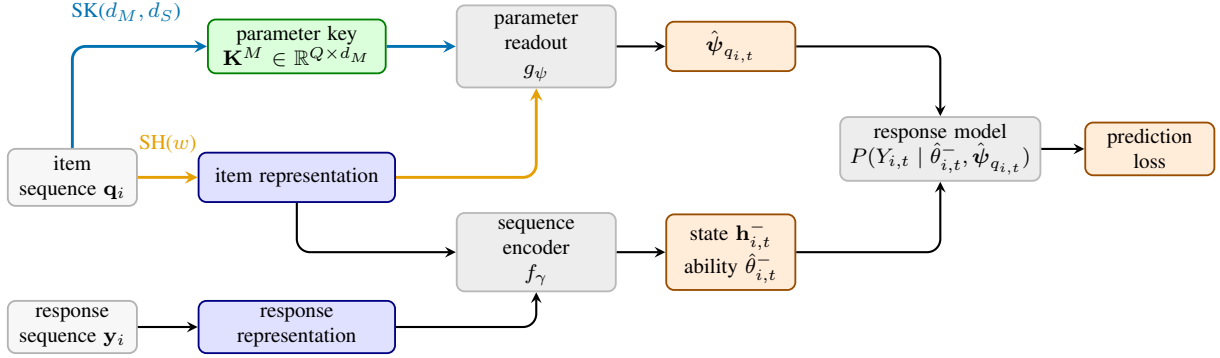
\begin{figure}[!htbp]
\centering
\Description{Two item representation paths feed a common response decoder.
In SH, one item table feeds both the sequence encoder and the item parameter
readout. In SK, a separate parameter key feeds only the parameter readout.
The sequence encoder uses item and response histories to estimate ability.
Ability and item parameters determine the predicted response and training loss.}
\resizebox{\linewidth}{!}{%
\begin{tikzpicture}[
    >=stealth, line width=0.7pt,
    box/.style={draw, rounded corners=3pt, minimum height=0.66cm,
                align=center, font=\scriptsize, inner sep=2.5pt},
    src/.style={box, fill=gray!6, draw=gray!60, text width=1.55cm},
    emb/.style={box, fill=blue!12, draw=blue!50!black,
                text width=2.45cm},
    key/.style={box, fill=green!14, draw=green!50!black,
                text width=2.20cm},
    mod/.style={box, fill=gray!14, draw=gray!55, text width=1.95cm},
    res/.style={box, fill=orange!15, draw=orange!60!black,
                text width=1.55cm},
    note/.style={font=\scriptsize, text=black!55},
    arr/.style={->, line width=0.8pt},
    sharedw/.style={->, line width=1.1pt, draw=SharedOrange},
    sepw/.style={->, line width=1.1pt, draw=SeparateBlue}
]
\definecolor{SharedOrange}{HTML}{E69F00}
\definecolor{SeparateBlue}{HTML}{0072B2}

\node[src] (q) at (0.85,1.0) {item\\sequence $\mathbf q_i$};
\node[src] (y) at (0.85,-1.0) {response\\sequence $\mathbf y_i$};

\node[key] (k) at (3.85,2.75)
  {parameter key\\$\mathbf K^M\in\mathbb R^{Q\times\dM}$};
\node[emb] (e) at (3.85,1.0)
  {item representation};
\node[emb] (r) at (3.85,-1.0) {response\\representation};

\draw[sharedw] (q) -- (e)
  node[midway, above=5pt, note, text=SharedOrange] {SH($w$)};
\draw[sepw, rounded corners=5pt] (q.north) |- (k.west)
  node[pos=0.70, above=5pt, note, text=SeparateBlue] {SK($\dM,\dS$)};
\draw[arr] (y) -- (r);

\node[mod] (enc) at (7.05,0.0) {sequence encoder\\$f_\gamma$};
\node[res] (th) at (9.65,0.0)
  {state $\mathbf h^-_{i,t}$\\ability $\hat\theta^-_{i,t}$};
\draw[arr, rounded corners=5pt] (e.south) |- (enc.west);
\draw[arr, rounded corners=5pt] (r.east) -| (enc.south);
\draw[arr] (enc) -- (th);

\node[mod] (g) at (7.05,2.75) {parameter readout\\$g_\psi$};
\node[res] (ab) at (9.65,2.75)
  {$\hat{\boldsymbol\psi}_{q_{i,t}}$};
\draw[sepw] (k) -- (g);
\draw[sharedw, rounded corners=5pt] (e.east) -| (g.south);
\draw[arr] (g) -- (ab);

\node[mod, text width=2.50cm] (response) at (12.45,1.38)
  {response model\\
   $P(Y_{i,t}\mid\hat\theta^-_{i,t},
   \hat{\boldsymbol\psi}_{q_{i,t}})$};
\node[res] (loss) at (15.25,1.38) {prediction\\loss};
\draw[arr, rounded corners=5pt] (th.east) -| (response.south);
\draw[arr, rounded corners=5pt] (ab.east) -| (response.north);
\draw[arr] (response) -- (loss);
\end{tikzpicture}
}
\caption{Item representation paths in SH and SK. The sequence encoder
summarizes interactions before $t$ as an ability estimate, while the parameter
readout estimates the parameter vector of the current item $q_{i,t}$. SH($w$)
supplies both paths from one item table of width $w$. SK($\dM,\dS$) uses a
parameter key of width $\dM$ for the item parameter readout and an item table
of width $\dS$ for the sequence encoder. The remaining components and the
prediction loss are unchanged.}
\label{fig:widths}
\end{figure}

To isolate these pathways, we compare the two item-representation designs in
Figure~\ref{fig:widths} across all nine combinations of three sequence encoders
(LSTM, Transformer, and DKVMN) and three response decoders (2PL, GPCM, and NRM).
The shared-head design, \(\mathrm{SH}(w)\), supplies a single item embedding
table of width \(w\) to both the sequence encoder and the item parameter
readout, coupling the sequence input width and parameter readout width at
\(w=\dM=\dS\). The separated-key design, \(\mathrm{SK}(\dM,\dS)\), assigns a
sequence item table of width \(\dS\) to the encoder and a separate parameter key
table of width \(\dM\) to the item parameter readout, while keeping the encoder
backbone, response equation, and training objective unchanged. Unlike Deep-IRT
\citep{yeung_deepirt_2019}, where the same item embedding queries DKVMN memory
and predicts item difficulty, the parameter key \(\mathbf K^M\) in
\(\mathrm{SK}(\dM,\dS)\) never enters the sequence encoder.

We combine controlled simulations, where known
generating parameters allow exact evaluation of recovery and item-specific
ability distortions under fixed and shuffled item orders, with real-data
benchmarks on TIMSS 2019 (GPCM) and EdNet-KT1 (2PL and NRM), where neural
estimates are compared against marginal maximum-likelihood (MML) reference
estimates. We first show that high recovery of difficulty, mean step
thresholds, and correct-option intercepts often coexists with much weaker
recovery of discrimination and the keyed option slope contrast, even under
shuffled item order. Under fixed item order, repeatable item-specific ability
offsets \(\phi_q\) and gains \(s_q\) emerge and degrade uncorrected parameter
recovery, whereas pairing a wide parameter key with a thin sequence representation in
\(\mathrm{SK}(64,8)\) reduces these fixed-order distortions and improves
cross-order rank agreement relative to both \(\mathrm{SH}(64)\) and
\(\mathrm{SK}(64,64)\).

Mapping the complete \((\dM,\dS)\) width grid shows that the two widths act in
opposite directions: widening \(\dM\) from 8 to 64 improves option-slope and
discrimination estimation across all nine simulated pairs and all nine \(\dS=8\) real-data pairs, whereas
widening \(\dS\) at \(\dM=64\) degrades simulated recovery by amplifying
item-specific ability distortions. Across the width grid, models
with nearly identical next-response test accuracy assign substantially
different rankings to item discrimination and option slopes. In simulation, and
in the majority of real-data comparisons, \(\mathrm{SK}(64,8)\) improves item
parameter estimation with fewer trainable parameters than \(\mathrm{SH}(64)\)
and the same number of stored item values as \(\mathrm{SK}(8,64)\), while
reversals on the EdNet Transformer show how narrowing
\(\dS\) to 8 can reduce predictive capacity on complex real interaction logs.

In summary, this study makes three contributions.
\begin{itemize}
\item \textbf{Analytical framework.} We derive the gradient asymmetries,
gradient-routing equations, and affine logit invariances that govern item
parameter estimation in end-to-end KT decoders across binary (2PL), polytomous
(GPCM), and nominal (NRM) response models (Section~\ref{sec:method}).
\item \textbf{Identification of order and width mechanisms.} Through controlled
manipulations of item order, independent width sweeps over \((\dM,\dS)\),
backward gradient blocking, and ability-readout ablations, we show why location
recovery exceeds slope recovery and why narrowing \(\dS\) while widening
\(\dM\) reduces item-order dependence (Sections~\ref{sec:results_trajectories}
and~\ref{sec:results_width}).
\item \textbf{Separating prediction from calibration.} Across 4,500 simulated runs per item order
and 1,420 real-data runs on TIMSS and EdNet (1,260 in the relative width plots), we demonstrate that test
accuracy and model size do not determine the quality of learned item parameters,
and specify the diagnostic checks required before interpreting IRT decoder
weights in KT (Sections~\ref{sec:results_accuracy}--\ref{sec:discussion}).
\end{itemize}

\section{Related Work}\label{sec:related}

\subsection{Interpretability in Knowledge Tracing}

Efforts to make neural KT models interpretable have targeted different stages of
the prediction pipeline. Standard deep KT architectures, including recurrent
models such as DKT \citep{piech_deep_2015}, key--value memory networks such as
DKVMN \citep{zhang_dkvmn_2017}, and attention-based models such as SAKT and AKT
\citep{pandey_sakt_2019,ghosh_akt_2020}, are trained and benchmarked primarily on
next-response prediction metrics such as AUC, accuracy, and cross-entropy loss
\citep{liu_pykt_2022,bai_survey_2024}. While post-hoc inspection of hidden
states, memory slots, or attention weights can illustrate how a prediction was
formed, those internal activations do not possess a fixed psychometric scale.

To obtain more structured representations, several KT architectures incorporate
explicit domain semantics into the encoder. Probabilistic and state-space
approaches such as IKT \citep{minn_ikt_2022} and PSI-KT \citep{zhou_psikt_2024}
model skill mastery, transfer profiles, cognitive traits, and prerequisite
graphs directly. Graph-based models such as GIKT \citep{yang_gikt_2020}
propagate question and skill embeddings over a question--skill bipartite graph
prior to prediction. A parallel line of work injects scalar difficulty terms or
question-level attributes into attention or state-update equations, as in AKT
\citep{ghosh_akt_2020}, simpleKT \citep{liu_simplekt_2023}, DIMKT
\citep{shen_dimkt_2022}, and DTransformer \citep{yin_dtransformer_2023}, while
Option Tracing extends prediction from binary correctness to the specific option
selected \citep{ghosh_option_2021}.

Although these designs attach domain labels to latent states, prerequisite
edges, or scalar difficulty modifiers, those quantities are not interchangeable
with the parameters of a formal IRT response model. An embedding scalar that
modulates attention weights or memory updates is defined relative to the rest of
the neural architecture rather than through an identified measurement equation.

\subsection{Neural Estimation of Item Response Models}

In classical psychometrics, IRT models specify the probability of a response as
a function of latent learner ability and item parameters, namely discrimination and
difficulty in the 2PL \citep{lord_1980}, discrimination and step
thresholds in the GPCM \citep{muraki_gpcm_1992}, and category-specific option
slopes and intercepts in the NRM \citep{bock_1972,thissen_steinberg_1986}.
Because ability and item parameters are estimated jointly from a static response
matrix (typically via marginal maximum likelihood \citep{bock_aitkin_1981}),
methodological studies evaluate parameter recovery against known
generating values in simulation alongside fit or missing-response prediction on
empirical data.

Recent neural IRT methods replace classical numerical integration with amortized
variational inference while retaining the exact IRT response likelihood.
ML2P-VAE \citep{curi_ml2pvae_2019,converse_estimation_2021} and VIBO
\citep{wu_variational_2020} train inference networks to map response vectors to
posterior distributions over ability and item parameters, and Variational
Temporal IRT extends this variational framework to time-varying proficiency
\citep{kim_variational_2023}. In particular, \citet{wu_variational_2020} observe that
unamortized variational inference can achieve superior item parameter recovery
in small samples while underperforming amortized inference on missing-response
prediction, providing direct evidence that predictive accuracy and parameter
recovery need not rank estimators identically.

However, amortized IRT inference differs structurally from sequential KT. In
sequential KT, the encoder updates a learner state step by step and receives
prediction gradients from subsequent interactions across the trajectory. When an
item representation participates in both encoding the interaction history and
reading out item parameters, sequential training couples state estimation with
item calibration in ways that static variational autoencoders do not encounter.

\subsection{Item Response Models as Knowledge Tracing Decoders}
\label{sec:irt_in_kt}

Placing an IRT response equation directly at the output of a sequential KT
encoder connects these two lines of work, but prior studies have examined only a restricted
subset of IRT parameters and evaluation criteria (Table~\ref{tab:related-outputs}).
Deep-IRT \citep{yeung_deepirt_2019} combines a DKVMN sequence backbone with a
one-parameter logistic (1PL) decoder whose discrimination is fixed at 3.0.
In that architecture, Deep-IRT uses the same item embedding \(\mathbf z_q\) both to query
the concept-key memory (forming the ability estimate) and to read out item
difficulty \(\hat\beta_q\). \citet{tsutsumi_independent_2021} observe that this
sharing allows current-item information to enter the ability estimate, and by
separating the student and item networks in a 1PL decoder, they raise simulated
difficulty correlation from 0.611 to 0.886. Neither study estimates item
discrimination, polytomous step thresholds, or nominal option slopes.

Subsequent KT studies invoke IRT structure primarily at the level of functional
form or additive scoring. QIKT \citep{chen_qikt_2023} sums question-specific
and general mastery scores inside a parameter-free sigmoid layer and evaluates
prediction accuracy rather than IRT parameter recovery. \citet{vie_kashima_2023}
demonstrate algebraically that the standard item-output DKT decoder is equivalent
to a dynamic multidimensional IRT model, but evaluate decoder choices solely
through test accuracy and AUC without comparing the learned item weights against
generating parameters or offline IRT calibrations. Conversely,
\citet{khajah_supercharging_2024} integrate multidimensional IRT with Bayesian
Knowledge Tracing (BKT) and show that higher predictive accuracy does not imply
better recovery of item-to-skill assignments, though the observation model
includes only item-specific guessing and slipping offsets rather than item
discrimination.

\begin{table}[!htbp]
\centering
\footnotesize
\caption{Item outputs and their evaluation in representative knowledge tracing
models that use item response terminology or an item response equation.}
\label{tab:related-outputs}
\renewcommand{\arraystretch}{1.06}
\begin{tabular}{@{}p{2.5cm}p{3.2cm}p{3.7cm}p{4.6cm}@{}}
\toprule
\RaggedRight Method &
\RaggedRight Item output &
\RaggedRight Use in response prediction &
\RaggedRight Evaluation of the item output \\
\midrule

\RaggedRight Deep-IRT \citep{yeung_deepirt_2019}
& \RaggedRight Item difficulty \(\beta_q\)
& \RaggedRight 1PL equation with discrimination fixed at 3.0
& \RaggedRight External real-data comparisons, with no simulated recovery of
\(\beta_q\) \\

\RaggedRight Independent-network Deep-IRT
\citep{tsutsumi_independent_2021}
& \RaggedRight Item difficulty and skill difficulty
& \RaggedRight 1PL equation with separate student and item networks
& \RaggedRight Simulated difficulty recovery and test prediction under fixed
discrimination \\

\RaggedRight QIKT \citep{chen_qikt_2023}
& \RaggedRight Question-specific acquisition/problem-solving and general mastery scores
& \RaggedRight Additive sigmoid over the three scores with no learnable
response-layer weights
& \RaggedRight Test prediction and component ablations, with no item parameter
recovery \\

\RaggedRight DKT as implicit dynamic MIRT \citep{vie_kashima_2023}
& \RaggedRight Item coefficient vector and intercept in item-output DKT
& \RaggedRight Inner product with a dynamic multidimensional learner state
& \RaggedRight Test accuracy and AUC across decoder choices, with no comparison
with generating or offline IRT parameters \\

\RaggedRight BKT with IRT and skill discovery
\citep{khajah_supercharging_2024}
& \RaggedRight Item-specific guessing/slipping offsets and item-to-skill
assignments
& \RaggedRight BKT observation probabilities with learner ability and item
offsets
& \RaggedRight BKT parameter and skill-assignment recovery, with no item
discrimination estimation \\

\RaggedRight Option Tracing \citep{ghosh_option_2021}
& \RaggedRight Selected option probabilities
& \RaggedRight Categorical prediction of the selected option
& \RaggedRight Test prediction, with no NRM option slope or intercept estimation
\\

\bottomrule
\end{tabular}
\end{table}

Prior sequential KT--IRT decoder studies thus either estimate only 1PL difficulty \citep{yeung_deepirt_2019,tsutsumi_independent_2021}
or evaluate 2PL and multidimensional IRT decoders solely through response
prediction \citep{vie_kashima_2023}.
Moving beyond 1PL difficulty to estimate 2PL discrimination, GPCM step
thresholds, and NRM option slopes introduces a structural difference between
location and slope identification in sequential models, which we
formalize next.

\section{Methodological Framework}\label{sec:method}

We formalize the SH and SK predictors and examine how next-response loss
updates their item parameters. The derivatives distinguish the parameter
groups, and the gradient paths distinguish the two item representations.
Affine transformations then connect changes in estimated ability to changes
in estimated item parameters. These results guide the tests of order
dependence, parameter recovery, test accuracy, and model size.

\subsection{Sequential Prediction and Item Representations}
\label{sec:prediction_representation}

The following notation formalizes the prediction paths in
Figure~\ref{fig:widths}.

For learner \(i\), let
\[
\mathcal H_{i,t-1}
=\{(q_{i,1},y_{i,1}),\ldots,(q_{i,t-1},y_{i,t-1})\}
\]
denote the interaction history before step \(t\), where \(q_{i,t}\) is the
current item and \(y_{i,t}\) is its response. For response set
\(\mathcal Y\), we use the standard next-response target of neural KT models
\citep{piech_deep_2015,zhang_dkvmn_2017,ghosh_akt_2020},

\begin{equation}
P\!\left(Y_{i,t}=k\mid q_{i,t},\mathcal H_{i,t-1}\right),
\qquad k\in\mathcal Y .
\label{eq:kt_target}
\end{equation}

A sequence encoder maps the history to
\(\mathbf h^-_{i,t}=f_{\gamma}(\mathcal H_{i,t-1})\). A readout gives the
scalar ability estimate
\(\hat\theta^-_{i,t}=r_{\omega}(\mathbf h^-_{i,t})\), where the superscript
\(-\) marks the state before \(y_{i,t}\) is observed. In IRT, ability denotes
the latent coordinate supplied to an item response model. Here that coordinate
is estimated from the KT history. It is neither observed nor trained against a
separate learner target. The current item enters through its estimated
parameter vector \(\hat{\boldsymbol\psi}_{q_{i,t}}\). The response model gives

\begin{equation}
\hat p_{i,t,k}
=P\!\left(
Y_{i,t}=k
\mid
\hat\theta^-_{i,t},
\hat{\boldsymbol\psi}_{q_{i,t}}
\right).
\label{eq:prediction}
\end{equation}

After prediction, \((q_{i,t},y_{i,t})\) enters
\(\mathcal H_{i,t}\). No ability or item parameter target is supplied during
training.

Items are represented by learned lookup vectors. At each previous step
\(s<t\), item \(q_{i,s}\) indexes an item vector and response \(y_{i,s}\)
indexes a response vector. The encoder processes these paired inputs to form
\(\mathbf h^-_{i,t}\). At the prediction step, the current item
\(q_{i,t}\) also indexes the vector supplied to \(g_\psi\), which estimates
the item parameters used in
Eq.~\eqref{eq:prediction}. Thus item identity enters two computations. Past
items contribute to the ability estimate through the sequence encoder, and
the current item contributes through the item parameter readout.

SH and SK differ only in whether these two computations use the same lookup
table. Let \(\mathbf E\in\mathbb R^{Q\times w}\) denote the item table in
SH(\(w\)). Its row \(\mathbf e_q\) represents item \(q\) in previous
interactions and is also supplied to \(g_\psi\) when \(q\) is the current
item. Let
\(\mathbf E^S\in\mathbb R^{Q\times\dS}\) and
\(\mathbf K^M\in\mathbb R^{Q\times\dM}\) denote the sequence table and
parameter key in SK(\(\dM,\dS\)), respectively. Previous items index rows
\(\mathbf e_q^S\) of \(\mathbf E^S\), while the current item indexes
\(\mathbf k_q^M\) for the parameter readout. The response representations
and each encoder's input operation remain unchanged. The parameter estimates
are

\begin{equation}
\hat{\boldsymbol\psi}_q
=
\begin{cases}
g_{\psi}(\mathbf e_q),
& \mathrm{SH}(w),\\[2pt]
g_{\psi}(\mathbf k_q^M),
& \mathrm{SK}(\dM,\dS),
\end{cases}
\qquad
\frac{\partial\mathbf h^-_{i,t}}{\partial\mathbf k_q^M}=0 .
\label{eq:sh_sk}
\end{equation}

Under SH, \(\mathbf e_q\) also enters \(f_{\gamma}\). Under SK,
\(\mathbf e_q^S\) enters \(f_{\gamma}\), while \(\mathbf k_q^M\) is used
only by \(g_{\psi}\). Thus \(\dS\) is the item input width of the sequence
encoder, and \(\dM\) is the input width of the item parameter readout.
SH(\(w\)) and SK(\(w,w\)) are not the same model. The former has one item
table and the latter has two.

\subsection{Item Response Models and Parameter Readouts}
\label{sec:response_models}

\paragraph{Item response models.}
We use the two-parameter logistic model (2PL) for binary responses
\citep{lord_1980,deayala_theory_2009}, the generalized partial credit model
(GPCM) for ordered responses \citep{muraki_gpcm_1992}, and the nominal
response model (NRM) for nominal responses \citep{thissen_steinberg_1986}.
Each model is used as the response
decoder in Eq.~\eqref{eq:prediction}. It maps the ability estimate and item
parameter readout to response probabilities. The item parameters are outputs of
\(g_\psi\), not separately observed training targets.

For a binary response, the 2PL uses discrimination \(\alpha_q>0\) and
difficulty \(\beta_q\),
\begin{equation}
P_{\mathrm{2PL}}(Y_{i,t}=1\mid\hat\theta^-_{i,t},q)
=\sigma\!\left[\alpha_q(\hat\theta^-_{i,t}-\beta_q)\right].
\label{eq:2pl}
\end{equation}
Difficulty is the ability value at which the predicted probability equals one
half, while discrimination controls how steeply the response probability changes
around that point.

For an ordered response with \(K\) categories, the GPCM uses one
discrimination \(\alpha_q>0\) and \(K-1\) step thresholds
\(\boldsymbol\beta_q=(\beta_{q,1},\ldots,\beta_{q,K-1})\),
\begin{equation}
P_{\mathrm{GPCM}}(Y_{i,t}=k\mid\hat\theta^-_{i,t},q)
=
\frac{
\exp\!\left[\sum_{m=1}^{k}
\alpha_q(\hat\theta^-_{i,t}-\beta_{q,m})\right]
}{
\sum_{j=0}^{K-1}\exp\!\left[\sum_{m=1}^{j}
\alpha_q(\hat\theta^-_{i,t}-\beta_{q,m})\right]
}.
\label{eq:gpcm}
\end{equation}
Here
\(\sum_{m=1}^{0}\alpha_q(\hat\theta^-_{i,t}-\beta_{q,m})=0\)
for category \(k=0\). A step threshold \(\beta_{q,m}\) is the ability value at
which adjacent categories \(m-1\) and \(m\) have equal probability.
Discrimination controls how rapidly their odds change around that threshold.

For a nominal response, the NRM assigns an option slope \(a_{q,k}\) and
intercept \(c_{q,k}\) to each option,
\begin{equation}
P_{\mathrm{NRM}}(Y_{i,t}=k\mid\hat\theta^-_{i,t},q)
=
\frac{\exp(a_{q,k}\hat\theta^-_{i,t}+c_{q,k})}
{\sum_{j=0}^{K-1}\exp(a_{q,j}\hat\theta^-_{i,t}+c_{q,j})}.
\label{eq:nrm}
\end{equation}
The option slope controls how the option logit changes with ability, while the
intercept gives its value at ability zero. Because only differences among
option logits affect the probability, the option slopes and intercepts are
centered within item.

The GPCM thresholds remain unsorted during training. When a comparison requires
one threshold value per item, we use their mean
\(\bar\beta_q=(K-1)^{-1}\sum_{m=1}^{K-1}\beta_{q,m}\). NRM option slopes are
identified through within-item constraints rather than an absolute slope level
\citep{bock_1972}. We therefore compare the correct option slope with the mean
slope of the other options. For correct option \(k_q^\star\), this keyed option
slope contrast is
\begin{equation}
a_q^{\mathrm{key}}
=a_{q,k_q^\star}
-\frac{1}{K-1}\sum_{k\ne k_q^\star}a_{q,k}.
\label{eq:nrm_key}
\end{equation}
A positive \(a_q^{\mathrm{key}}\) means that the correct option logit increases
with ability faster than the mean logit of the other options. Under Bock's
sum-to-zero constraint, \(a_q^{\mathrm{key}}=K a_{q,k_q^\star}/(K-1)\), so the
contrast is a rescaled correct option slope rather than an additional NRM
parameter. NRM option slope recovery and agreement are evaluated using
\(a_q^{\mathrm{key}}\). Correlations pooled over all item and option pairs are
excluded because they combine within-item option differences with between-item
differences.

\paragraph{Parameter readouts.}
For the following readouts, \(\mathbf x_q=\mathbf e_q\) under SH and
\(\mathbf x_q=\mathbf k_q^M\) under SK. For the 2PL and GPCM,

\begin{align}
\alpha_q
&=\operatorname{softplus}
  (\mathbf w_\alpha^\top\mathbf x_q+b_\alpha),
&
\boldsymbol\beta_q
&=W_\beta\mathbf x_q+\mathbf b_\beta .
\label{eq:ordered_readout}
\end{align}

For the 2PL, the vector \(\boldsymbol\beta_q\) reduces to the scalar \(\beta_q\), whereas for the GPCM it contains the
\(K-1\) thresholds.

For the NRM, to parameterize the keyed option slope contrast
\(a_q^{\mathrm{key}}\) in Eq.~\eqref{eq:nrm_key} directly while enforcing
within-item centering, define \(\mathbf d_q\in\mathbb R^K\) to take value
\((K-1)/K\) at \(k_q^\star\) and \(-1/K\) elsewhere, let
\(C=I-K^{-1}\mathbf 1\mathbf 1^\top\) be the centering matrix, and let
\(\Pi_q^\perp\) project a centered vector onto the subspace orthogonal to
\(\mathbf d_q\). The NRM readout is

\begin{align}
\kappa_q
&=\operatorname{softplus}
  (\mathbf w_\kappa^\top\mathbf x_q+b_\kappa),
&
\mathbf a_q
&=\kappa_q\mathbf d_q
  +\Pi_q^\perp(W_a\mathbf x_q+\mathbf b_a),
\nonumber\\
\mathbf c_q
&=C(W_c\mathbf x_q+\mathbf b_c).
\label{eq:nrm_readout}
\end{align}

Equation~\eqref{eq:nrm_readout} gives
\(\kappa_q=a_q^{\mathrm{key}}\ge0\). Consequently, an item with negative
generating \(a_q^{\mathrm{key}}\) cannot be represented by this readout.

Let
\(\mathbf a_q^\perp=\Pi_q^\perp(W_a\mathbf x_q+\mathbf b_a)\).
The NRM logits used in training are

\begin{equation}
z^{\mathrm{route}}_{i,t,k}
=\kappa_qd_{q,k}\hat\theta^-_{i,t}
+a^\perp_{q,k}\sg[\hat\theta^-_{i,t}]
+c_{q,k}.
\label{eq:nrm_route}
\end{equation}

Equation~\eqref{eq:nrm_route} has the same forward value as
Eq.~\eqref{eq:nrm}. The stop gradient changes only the derivative passed to
\(\hat\theta^-_{i,t}\). The derivatives used to estimate
\(\mathbf a_q\) and \(\mathbf c_q\) are unchanged. SH and SK use the same
NRM routing.

Notice that this readout structure also distinguishes SK from Deep-IRT
\citep{yeung_deepirt_2019} when a DKVMN encoder is used. Whereas Deep-IRT uses a
single item embedding \(\mathbf z_q\) both to address concept-key memory and to
read out item difficulty, SK supplies \(\mathbf k_q^M\) exclusively to
\(g_\psi\) and uses \(\mathbf e_q^S\) inside the sequence encoder (see
Section~\ref{sec:affine_invariance} for the resulting gradient decomposition).

All components are trained end to end with next-response loss. The 2PL
uses binary cross-entropy, the GPCM uses cumulative ordinal cross-entropy, and
the NRM uses categorical cross-entropy. SH and SK use the same response
function, target, and loss. The loss definitions and optimization settings are
given in Section~\ref{sec:experimental_setup}.

\subsection{Parameter Gradients and Fisher Information}\label{sec:item_estimation}
Because the three response models are differentiable, the next-response loss
provides an explicit gradient for every item parameter. These gradients connect
the response-model definitions in Section~\ref{sec:response_models} to how the
KT model updates them during
training. For one 2PL response, let
\(
p=\sigma[\alpha_q(\hat\theta^-_{i,t}-\beta_q)]
\). The derivatives of binary cross-entropy are

\begin{align}
\frac{\partial\mathcal L_{i,t}}{\partial\alpha_q}
&=(\hat\theta^-_{i,t}-\beta_q)(p-y_{i,t}),
&
\frac{\partial\mathcal L_{i,t}}{\partial\beta_q}
&=-\alpha_q(p-y_{i,t}).
\label{eq:2pl_gradients}
\end{align}

For the GPCM, define

\begin{equation}
R_{i,t,m}
=\frac{\partial\mathcal L_{i,t}}
{\partial[\alpha_q(\hat\theta^-_{i,t}-\beta_{q,m})]} .
\label{eq:gpcm_threshold_derivative}
\end{equation}

Then

\begin{align}
\frac{\partial\mathcal L_{i,t}}{\partial\alpha_q}
&=\sum_{m=1}^{K-1}
(\hat\theta^-_{i,t}-\beta_{q,m})R_{i,t,m},
&
\frac{\partial\mathcal L_{i,t}}{\partial\beta_{q,m}}
&=-\alpha_qR_{i,t,m}.
\label{eq:gpcm_gradients}
\end{align}

For the NRM, let
\(
G_{i,t,k}
=p_{i,t,k}-\mathbb I\{y_{i,t}=k\}
\). Before applying the centering projections,

\begin{align}
\frac{\partial\mathcal L_{i,t}}{\partial a_{q,k}}
&=\hat\theta^-_{i,t}G_{i,t,k},
&
\frac{\partial\mathcal L_{i,t}}{\partial c_{q,k}}
&=G_{i,t,k}.
\label{eq:nrm_gradients}
\end{align}

The 2PL discrimination derivative in Eq.~\eqref{eq:2pl_gradients} multiplies
the prediction residual by the ability-to-difficulty distance. Each term in
the GPCM discrimination derivative in Eq.~\eqref{eq:gpcm_gradients} is weighted
by an ability-to-threshold distance. The NRM option slope derivative in
Eq.~\eqref{eq:nrm_gradients} multiplies the category residual by
\(\hat\theta^-_{i,t}\). The derivatives for difficulty, thresholds, and
option intercepts contain neither the ability estimate nor an
ability-to-location distance outside the residual term.

For the 2PL, the conditional Fisher information from one response
\citep{lord_1980,baker_kim_2004} includes

\begin{align}
I_{\alpha\alpha}
&=(\hat\theta^-_{i,t}-\beta_q)^2p(1-p),
&
I_{\beta\beta}
&=\alpha_q^2p(1-p).
\label{eq:2pl_information}
\end{align}

Here Fisher information is the expected curvature of the negative conditional
response log-likelihood with respect to each item parameter.

At \(\hat\theta^-_{i,t}=\beta_q\),
\(I_{\alpha\alpha}=0\), whereas \(I_{\beta\beta}>0\) whenever
\(\alpha_q>0\) and \(0<p<1\).
Thus a response with \(\hat\theta^-_{i,t}\) near \(\beta_q\) can update
difficulty while providing little first-order information about \(\alpha_q\). The GPCM and NRM
derivatives show that discrimination and option slope updates also depend on
the distribution of the ability estimates. These expressions do not predict
the recovery difference between discrimination or option slopes and the
location or intercept parameters. Section~\ref{sec:results} measures that
difference.

Intuitively, an ability estimate near an item's difficulty can help locate the
response curve while saying little about how sharply it changes with ability. In the
NRM, an ability estimate near zero likewise reduces the update to the option
slopes but not to the option intercepts. This is why the results report
discrimination and option slope recovery separately from difficulty,
threshold, and intercept recovery.

\subsection{Gradient Routing and Affine Logit Invariance}\label{sec:affine_invariance}

\paragraph{Gradient routing.}
Let
\(\mathcal O_q=\{(i,t):q_{i,t}=q\}\), and define

\begin{equation}
\mathbf u_q
=\sum_{(i,t)\in\mathcal O_q}
\frac{\partial\mathcal L_{i,t}}
{\partial\hat{\boldsymbol\psi}_q},
\qquad
J_q=\frac{\partial\hat{\boldsymbol\psi}_q}
{\partial\mathbf x_q}.
\label{eq:item_derivative}
\end{equation}

For the NRM, \(\widetilde\nabla\) denotes the derivative after the
stop gradient in Eq.~\eqref{eq:nrm_route}. For the 2PL and GPCM,
\(\widetilde\nabla=\nabla\).

The derivative with respect to an item lookup vector differs between SH and
SK. Define
\begin{equation}
D_q
=\sum_{i,t}
\widetilde\nabla_{\mathbf h^-_{i,t}}L_{\mathrm{pred}}
\frac{\partial\mathbf h^-_{i,t}}{\partial\mathbf e_q}.
\label{eq:sequence_gradient}
\end{equation}
The gradients for the SH and SK item tables are
\begin{align}
\widetilde\nabla_{\mathbf e_q}L_{\mathrm{pred}}
&=J_q^\top\mathbf u_q+D_q,
&& \mathrm{SH}(w),
\label{eq:sh_gradient}\\
\widetilde\nabla_{\mathbf k_q^M}L_{\mathrm{pred}}
&=J_q^\top\mathbf u_q,
&
\widetilde\nabla_{\mathbf e_q^S}L_{\mathrm{pred}}
&=D_q,
&& \mathrm{SK}(\dM,\dS).
\label{eq:sk_gradient}
\end{align}
Under SH, \(\mathbf e_q\) is an input to both \(g_\psi\) and
\(f_\gamma\), so the chain rule gives the two terms in
Eq.~\eqref{eq:sh_gradient}. Under SK,
\(\partial\mathbf h^-_{i,t}/\partial\mathbf k_q^M=0\), so the two terms
apply to different tables.

If the SK parameter key block is stationary and \(J_q\) has full row rank
after the response model constraints, Eq.~\eqref{eq:sk_gradient} implies
\(\mathbf u_q=0\). For the routed NRM, this rank condition can fail at the
softplus floor \(a_q^{\mathrm{key}}=0\). Under SH, stationarity requires \(J_q^\top\mathbf u_q+D_q=0\), meaning
that parameter-readout gradients \(J_q^\top\mathbf u_q\) and sequence-encoder
gradients \(D_q\) are summed on the same item vector throughout training.

\paragraph{Affine invariance of response logits.}
The gradient paths describe how the item representations are updated. A
separate question is whether the response probabilities distinguish changes
in estimated ability from changes in estimated item parameters.
A sequential decoder can share the response equation of an IRT model
\citep{vie_kashima_2023}, but that equation does not determine the scale of
the resulting ability and item parameter estimates. IRT ability and item
parameters are defined relative to a common scale, and an affine change of that
scale induces corresponding changes in the item parameters
\citep{lord_1980,baker_kim_2004}. A common positive affine change preserves
the ordering of discrimination and difficulty across items. It cannot explain
differences in their rank recovery. Item-specific changes require an additional
condition on the ability estimates supplied to the decoder. For response logit vector
\(\mathbf z_{i,t}\),

\begin{equation}
d\mathbf z_{i,t}
=J_{\theta}\,d\hat\theta^-_{i,t}
+J_{\psi}\,d\hat{\boldsymbol\psi}_q .
\label{eq:logit_differential}
\end{equation}

Backpropagation multiplies the prediction residual by
\(J_{\theta}^{\top}\) and \(J_{\psi}^{\top}\). Joint changes in the
ability estimate and item parameters satisfying

\begin{equation}
J_{\theta}\,d\hat\theta^-_{i,t}
+J_{\psi}\,d\hat{\boldsymbol\psi}_q=0
\label{eq:null_direction}
\end{equation}

leave the logits unchanged to first order. For the 2PL, one such direction is

\begin{equation}
d\theta=u_q+r_q\theta,\qquad
d\alpha_q=-r_q\alpha_q,\qquad
d\beta_q=u_q+r_q\beta_q .
\label{eq:2pl_null_tangent}
\end{equation}

Equation~\eqref{eq:null_direction} is also a null direction of the
prediction loss gradient. If
\(\mathbf v=(d\hat\theta^-_{i,t},
d\hat{\boldsymbol\psi}_q)\) satisfies the equation, then
\(\mathbf v^\top\nabla\mathcal L_{i,t}=0\). Prediction loss therefore
cannot distinguish parameters along this direction to first order.

Suppose that, for responses to item \(q\), the ability estimate and a
reference ability satisfy
\begin{equation}
\hat\theta^-_{i,t}=s_q\theta^{\mathrm{ref}}_{i,t}+\phi_q,
\qquad s_q>0.
\label{eq:state_affine}
\end{equation}
Here \(\phi_q\) is the additive offset associated with item \(q\), and \(s_q\)
is the multiplicative gain relative to the reference ability. A gain of one
preserves the reference scale, a gain above one expands it, and a gain between
zero and one compresses it.
The 2PL and GPCM logits are unchanged on the reference scale under
\begin{equation}
\alpha_q^{\mathrm{ref}}=\hat\alpha_qs_q,
\qquad
\beta_{q,m}^{\mathrm{ref}}
=\frac{\hat\beta_{q,m}-\phi_q}{s_q}.
\label{eq:ordered_correction}
\end{equation}
The NRM logits are unchanged under
\begin{equation}
a_{q,k}^{\mathrm{ref}}=\hat a_{q,k}s_q,
\qquad
c_{q,k}^{\mathrm{ref}}
=\hat c_{q,k}+\hat a_{q,k}\phi_q.
\label{eq:nrm_correction}
\end{equation}
These transformations follow directly from the response logits. For the 2PL
and each GPCM threshold,
\[
\hat\alpha_q
\left(\hat\theta^-_{i,t}-\hat\beta_{q,m}\right)
=
\hat\alpha_qs_q
\left(
\theta^{\mathrm{ref}}_{i,t}
-\frac{\hat\beta_{q,m}-\phi_q}{s_q}
\right).
\]
For the NRM,
\[
\hat a_{q,k}\hat\theta^-_{i,t}+\hat c_{q,k}
=(\hat a_{q,k}s_q)\theta^{\mathrm{ref}}_{i,t}
+(\hat c_{q,k}+\hat a_{q,k}\phi_q).
\]

Equation~\eqref{eq:nrm_correction} preserves the NRM constraints because
\(\sum_k\hat a_{q,k}=\sum_k\hat c_{q,k}=0\). The nonnegative
\(a_q^{\mathrm{key}}\) constraint is also preserved for \(s_q>0\).
These are decoder-level identities. They do not establish that any
item-specific transformation can be implemented by a fixed sequence encoder.

The finite transformation and the gradient null direction describe the same
response equivalence. To express the inverse change of coordinates, set
\(s_q(\epsilon)=1+\epsilon r_q\) and
\(\phi_q(\epsilon)=\epsilon u_q\), with
\(\theta'=s_q\theta+\phi_q\), \(\alpha'_q=\alpha_q/s_q\), and
\(\beta'_{q,m}=s_q\beta_{q,m}+\phi_q\).
Differentiating at \(\epsilon=0\) gives
Eq.~\eqref{eq:2pl_null_tangent}. For the NRM, the corresponding change is
\(a'_{q,k}=a_{q,k}/s_q\) and
\(c'_{q,k}=c_{q,k}-a_{q,k}\phi_q/s_q\), giving
\(d a_{q,k}=-r_q a_{q,k}\) and
\(d c_{q,k}=-u_q a_{q,k}\). Thus the first-order null direction of the
prediction loss is tangent to the same affine transformation used to correct
the item-parameter estimates in
Section~\ref{sec:results_trajectories}.

We estimate \(\phi_q\) and \(s_q\) from the stored ability estimates before
each response. Let \(S_{i,t}=\hat\theta^-_{i,t}\) denote the estimate paired
with \(q_{i,t}\). We remove the learner mean and sequence position mean as

\begin{equation}
R_{i,t}
=S_{i,t}-\bar S_{i,\cdot}-\bar S_{\cdot,t}+\bar S_{\cdot,\cdot}.
\label{eq:state_residual}
\end{equation}

For items with at least 20 paired responses,

\begin{equation}
\phi_q
=\frac{1}{|\mathcal O_q|}
\sum_{(i,t)\in\mathcal O_q}R_{i,t}.
\label{eq:state_offset}
\end{equation}

The gain \(s_q\) is the ordinary least-squares coefficient from regressing
\(S_{i,t}\) on reference ability over \(\mathcal O_q\), with an intercept.
The reference is generating ability in simulation and the expected posterior
ability estimate from the offline IRT model for real data.

The demeaned offset in Eq.~\eqref{eq:state_offset} need not equal the
regression intercept. Nor does a fitted affine relation imply exact response
equivalence. We therefore fit a second offset \(\tilde\phi_q\) jointly with
\(s_q\) and retain the residual,
\begin{equation}
S_{i,t}=s_q\theta^{\mathrm{ref}}_{i,t}+\tilde\phi_q+\varepsilon_{i,t}.
\label{eq:affine_residual}
\end{equation}
After the parameter transformation, the remaining 2PL logit difference is
\(\hat\alpha_q\varepsilon_{i,t}\). For the GPCM it is
\(k\hat\alpha_q\varepsilon_{i,t}\) at category \(k\), and for the NRM it is
\(\hat a_{q,k}\varepsilon_{i,t}\). These residuals measure how closely the
fitted affine relation accounts for the trained model's logits on responses
not used to estimate the transformation.

Equations~\eqref{eq:sh_gradient} and \eqref{eq:sk_gradient} specify how the
prediction residual updates the item representations while joint changes in
ability and item parameters can preserve the response probabilities. Under
SH, \(D_q\) and \(J_q^\top\mathbf u_q\) update the same table, so one update
can change both \(\hat\theta^-_{i,t}\) and the item parameters. Under SK, the
parameter key cannot change \(\hat\theta^-_{i,t}\), because \(D_q\) updates
only \(\mathbf E^S\). This prevents the parameter key from directly
changing both terms in a response probability, but it does not fix the scale
of the ability estimates produced by the sequence encoder. The affine
comparison measures item-specific offsets, gains, and the unexplained
variation in those ability estimates. It does not assume exact parameter recovery under SK.

Because \(\hat\theta^-_{i,t}\) is computed from interactions before \(t\)
and does not read \(q_{i,t}\) directly, differences in \(\phi_q\) and \(s_q\)
across items require the preceding history to carry information about the
current item.
The fixed order tiles one random permutation of the 200 items and assigns each
learner a consecutive segment of 60 items. Each item therefore appears at
three of the 60 sequence positions, while its successor in the tiled
permutation is fixed. The shuffled order condition replaces the tiled
permutation with an independently sampled and ordered item set for each learner.
It preserves \(N\), \(Q\), \(T\), the generating distributions, and the
response model, but not the exact exposure count for each item.
The residual in Eq.~\eqref{eq:state_residual} removes a mean associated
only with sequence position. It does not remove an interaction between an
item and the three positions that item occupies. The shuffled order condition
is therefore needed to test whether the item-specific values of \(\phi_q\) and
\(s_q\) remain when items visit all sequence positions.

These derivations motivate three empirical comparisons. The parameter
gradients motivate evaluating discrimination and option slopes separately
from difficulty, thresholds, and intercepts. The affine relations call for
testing whether item order changes ability offsets, gains, and item parameter
estimates. The gradient paths point to comparing SH and SK and varying
\(\dM\) and \(\dS\) independently. None of the derivations establishes a
recovery advantage for SK or a preferred width pair. Section~\ref{sec:results}
tests these comparisons and whether any gain in recovery or offline agreement
requires a loss of test accuracy or an increase in model size.

\subsection{Experimental Setup}\label{sec:experimental_setup}

\paragraph{Datasets.}
The simulated data contain \(N=2000\) learners and \(Q=200\) items. Each
learner answers \(T=60\) distinct items, giving 120,000 responses per
generated dataset. We generate five datasets,
indexed by \(d\in\{1,\ldots,5\}\), and divide each into five learner folds,
indexed by \(f\in\{1,\ldots,5\}\). Four folds form the training split and
one fold forms the test split. Binary, ordered, and nominal responses are
generated from the 2PL, GPCM, and NRM, respectively. Each learner's generating
ability follows \(\mathcal N(0,1)\) and remains fixed across the sequence,
providing a baseline in which any temporal or
item-order variation in \(\hat\theta^-_{i,t}\) arises from the estimator rather
than from non-stationary true ability. For
the 2PL and GPCM, \(\log\alpha_q\sim\mathcal N(0,0.3^2)\). The 2PL
difficulties and the unsorted GPCM step thresholds follow \(\mathcal N(0,1)\).
The thresholds are then ordered within each item. For the NRM, a latent
correct-option location follows \(\mathcal N(0,1)\). Before within-item
centering, the correct-option slope is lognormal with log standard deviation
0.3, and its intercept is the negative product of slope and location.
Distractor slopes and intercepts follow \(\mathcal N(0,0.6^2)\). The NRM
slopes and intercepts are then centered within each item. The fixed order
condition tiles one random permutation of the item bank and assigns each
learner a consecutive 60-item segment. It gives exactly 600 responses per
item and a deterministic next item map. The shuffled order condition samples
and orders 60 distinct items independently for each learner. Across the
stored shuffled order datasets, item exposure ranges from 525 to 661.

The real-data studies use TIMSS 2019 United States Grade 8 mathematics and
EdNet-KT1 \citep{choi_ednet_2020}. The retained TIMSS data contain 5,135
learners, 31 partial-credit items, and 28,324 responses in
\(\{0,1,2\}\). The response model is the GPCM. Item administration times are
unavailable, so responses are ordered by item identifier within learner. This
ordering is not chronological. We therefore use TIMSS for item parameter
comparisons and do not use it to evaluate sequence dynamics.

EdNet contains more than 250 items, many with few recorded responses. We retain
the 250 four-option items with the largest response counts. The resulting
subset contains 8,493 learners and 381,848 responses. The 2PL study uses
correctness, and the NRM study uses the selected option. Both studies use
the same interaction histories.

\paragraph{Implementation details.}
We evaluate LSTM, Transformer, and DKVMN sequence encoders with the 2PL,
GPCM, and NRM response models. All sequence encoders use hidden width 32.
The item representation widths are \(\{8,16,32,64\}\). For each encoder and
response model pair, SH is trained at the four shared widths and SK
is trained on the complete \(4\times4\) grid of \((\dM,\dS)\), giving 20
width choices.

For each item order, the simulated experiment contains 180 combinations of
three encoders, three response models, and 20 SH or SK width choices. Every
combination contains 25 training runs from five generated datasets and five
test folds, giving 4,500 runs per item order. TIMSS with GPCM, EdNet with 2PL,
and EdNet with NRM use the same 20 width choices with each encoder. Among these
180 real-data combinations, 26 contain 25 training runs and 154 contain five (1,420 runs in total).
Figures retain the individual runs so that cells with 25 runs are visibly
denser than cells with five runs.

Training uses Adam for 120 epochs with learning rate \(10^{-2}\), no weight
decay, no gradient clipping, and no early stopping. Simulated runs use full
batches. Real data runs use mini batches of 512 learners. Within each item-order
condition, paired architecture comparisons use the same responses, train/test split, and initialization
index. Generating item parameters and the offline IRT estimates are used only
for evaluation.

The 2PL uses binary cross-entropy and the NRM uses categorical cross-entropy.
For the GPCM, define
\(\hat F_{i,t,m}=\sum_{k=m}^{K-1}\hat p_{i,t,k}\) and
\(\nu_{i,t,m}=\mathbb I\{y_{i,t}\ge m\}\). Its cumulative ordinal
cross-entropy is
\begin{equation}
\mathcal L^{\mathrm{GPCM}}_{i,t}
=-\sum_{m=1}^{K-1}
\left[
\nu_{i,t,m}\log\hat F_{i,t,m}
+(1-\nu_{i,t,m})\log(1-\hat F_{i,t,m})
\right].
\label{eq:gpcm_loss}
\end{equation}
At \(K=2\), this loss reduces to binary cross-entropy. For the selected
response model, training minimizes the sum of event losses over the training
responses.

For the later response gradient experiment, the sequence encoder receives the
SK sequence item representation together with a linear read of the parameter
key. In the unblocked condition, gradients from later prediction losses pass
through this read into the parameter key. The blocked condition applies a stop gradient at
the read. From identical weights, the two conditions produce the
same forward values and differ only in the gradient passed from the sequence
encoder to the parameter key. We set
\(\dM=64\), vary \(\dS\) over \(\{8,16,32,64\}\), and repeat the comparison
for every encoder and response model using five generated datasets and five
test folds. A separate DKVMN experiment compares the same two conditions at
width 8 using the read-before-update order of the
original DKVMN model. It includes 25 paired runs for each response model.

For the affine channel analysis (Figure~\ref{fig:channels-2pl} and
Appendix~\ref{app:order-comparison}), the left and middle columns use
\(\mathrm{SK}(8,8)\) and the right column varies \(\dM\) at \(\dS=8\). Within
each generated dataset, we average \(\phi_q\), \(s_q\), and each item parameter
estimate over the 15 runs sharing the same generating parameters (three encoders
across five test folds), pair these averages with the generating values, and pool
the 200 item pairs across the five datasets. Reported slopes are ordinary
least-squares coefficients with an intercept (using the log scale for 2PL and
GPCM discrimination and the native scale for \(a_q^{\mathrm{key}}\)). To test
whether table separation or sequence width drives order sensitivity, we compare
\(\mathrm{SH}(64)\), \(\mathrm{SK}(64,64)\), and \(\mathrm{SK}(64,8)\) under
matched fixed and shuffled administrations, evaluating both ability--parameter
associations and cross-order item-rank agreement, while independent-learner checks of
the affine approximation appear in Appendix~\ref{app:affine-checks}. Finally, a
supplementary DKVMN readout ablation (detailed in
Appendix~\ref{app:dkvmn-readout}) tests whether passing the current item key
directly into the ability readout induces ability--difficulty correlations even under
random item order.

\paragraph{Evaluation metrics.}
\begin{itemize}
\item \textbf{Test accuracy (\(\mathrm{Acc}_{\mathrm{test}}\)).} We score
the final ten valid positions of each
test sequence. A prediction is correct when the response category with the
largest predicted probability equals the observed response. This is binary
correctness for the 2PL and multiclass accuracy for the GPCM and NRM. No
accuracy threshold is imposed by itself. Paired changes are assessed by the
simulated-data rule in Eq.~\eqref{eq:decision_rule}.

\item \textbf{Test log loss (\(\mathrm{NLL}_{\mathrm{test}}\)).}
We average the negative log probability of the observed response over the
same scored test positions. Unlike accuracy, log loss also
evaluates the probabilities assigned to correct and incorrect predictions.
We report paired changes in Appendix~\ref{app:probability-checks}, without
applying the accuracy threshold of 0.05 to this different scale.

\item \textbf{Simulated item parameter recovery
(\(\rho_{\xi}^{\mathrm{sim}}\)).} For item parameter \(\xi_q\),
\(\rho_{\xi}^{\mathrm{sim}}
=\rho_{\mathrm S}(\{\hat\xi_q\},\{\xi_q^\star\})\) is the Spearman rank
correlation between the estimate and its generating value. The statistic
measures whether the model preserves the ordering of items on that parameter.
Larger values indicate stronger recovery of the true item ordering. We report
discrimination \(\alpha_q\) and difficulty \(\beta_q\) for the 2PL,
discrimination, mean step threshold \(\bar\beta_q\), and threshold recovery for
the GPCM, and the keyed option slope contrast \(a_q^{\mathrm{key}}\) and
correct option intercept \(c_{q,k_q^\star}\) for the NRM, both before and after
the affine corrections in Eqs.~\eqref{eq:ordered_correction}
and~\eqref{eq:nrm_correction}.

\item \textbf{Agreement with an offline IRT reference
(\(\rho_{\xi}^{\mathrm{MML}}\)).} On TIMSS and EdNet,
generating parameters are unavailable. We therefore report
\(\rho_{\xi}^{\mathrm{MML}}\), the Spearman rank correlation between the
neural estimate \(\hat\xi_q\) and the corresponding offline
estimate obtained by fitting a one-dimensional
2PL, GPCM, or NRM by marginal maximum likelihood \citep{bock_aitkin_1981} with
the \texttt{mirt} package \citep{chalmers_mirt_2012} on the same responses.
This evaluates agreement with a separately
estimated parameter ordering, not recovery of ground truth. All 250 EdNet items enter the 2PL comparison, and the 186 items with finite
offline option slope estimates enter the NRM comparison.

\item \textbf{Agreement across item orders
(\(\rho_{\xi}^{\mathrm{ord}}\)).} We compute
\(\rho_{\xi}^{\mathrm{ord}}=
\rho_{\mathrm S}(\{\hat\xi_q^{\mathrm{fixed}}\},
\{\hat\xi_q^{\mathrm{shuffled}}\})\) within each matched dataset and fold.
Larger values indicate less dependence of the estimated item ranking on
administration. Unlike recovery, this comparison uses two learned estimates
of the same item parameter rather than the generating value. It uses all
200 items and is reported separately for each parameter group.

\item \textbf{Change from SH(8)
(\(\Delta_{\mathrm{SH}(8)}\rho_\xi\)).} In the relative width and trainable
parameter count figures, \(\rho_\xi\) denotes
\(\rho_\xi^{\mathrm{sim}}\) for simulated data and
\(\rho_\xi^{\mathrm{MML}}\) for real data. For model \(A\), we define
\(\Delta_{\mathrm{SH}(8)}\rho_\xi(A)
=\rho_\xi(A)-\rho_\xi(\mathrm{SH}(8))\), using SH(8) with the same dataset,
encoder, and response model. SH(8) is therefore zero, and a positive value
denotes stronger recovery or agreement than SH(8).

\item \textbf{Item-specific ability offset and gain (\(\phi_q\) and \(s_q\)).} The
offset \(\phi_q\) and gain \(s_q\) for item \(q\) are defined in
Eqs.~\eqref{eq:state_offset} and \eqref{eq:state_affine}. We report their Spearman correlations with the generating item parameters
(where values near zero indicate weak rank association with the generating parameters),
along with the split-half repeatability of \(\phi_q\) across disjoint halves of
the test set and item parameter recovery before and after affine correction.

\item \textbf{Model size (\(P\)).} \(P\) is the number of trainable parameters.
For a matched comparison, we report both the count and the ratio of the SK
count to the SH count. This tests whether a difference in item parameter
recovery can be attributed to a larger trained model.
\end{itemize}

\paragraph{Paired comparisons and decision rule.}
For any metric \(m\), the paired change from model \(B\) to model \(A\) is
\(\Delta_m(A,B)=m(A)-m(B)\). The complete width grid is reported, while the
primary comparisons set \(A=\mathrm{SK}(64,8)\) and compare it with
\(B\in\{\mathrm{SH}(8),\mathrm{SH}(64),\mathrm{SK}(8,64)\}\).
SK(64,8) and SK(8,64) store the same number of item table values. Their total
trainable parameter counts can differ because \(\dS\) changes the sequence
encoder input, so those counts are reported rather than assumed equal.

For simulated data, paired changes are first averaged over the five test
folds within each generated dataset and then over the five generated
datasets. Thus \(\bar\Delta_d=5^{-1}\sum_f\Delta_{d,f}\) and
\(\bar\Delta=5^{-1}\sum_d\bar\Delta_d\). Confidence intervals resample the
five generated datasets while
retaining all five folds within each resampled dataset. The predeclared
simulated data decision rule is

\begin{equation}
\bar\Delta\ge0.05
\quad\text{and}\quad
t_{\mathrm{paired}}
=\frac{\bar\Delta}{s_{\bar\Delta}/\sqrt{5}}\ge2.78 ,
\label{eq:decision_rule}
\end{equation}

where \(s_{\bar\Delta}\) is the sample standard deviation of
\(\{\bar\Delta_d\}_{d=1}^{5}\). This rule is not applied to real data.
Real-data figures instead report every training run together with the
corresponding mean for that width choice. Agreement across training runs with
different folds and initializations is reported in the Appendix. It is not
included in the decision rule.

\section{Results}\label{sec:results}

All models are trained solely with next-response loss, while generating item
parameters (in simulation) and offline IRT estimates (on real data) are used
exclusively for evaluation. Guided by Sections~\ref{sec:item_estimation}
and~\ref{sec:affine_invariance}, our evaluation proceeds in four steps.
Section~\ref{sec:results_trajectories} compares parameter recovery across
location and slope/discrimination groups under fixed and shuffled item orders,
connecting fixed-order estimation errors to item-specific ability offsets and
gains and testing whether \(\mathrm{SK}(64,8)\) reduces them.
Section~\ref{sec:results_width} then varies \(\dM\) and \(\dS\) across the
complete width grid to separate the effect of parameter key width from that of
sequence input width. Section~\ref{sec:results_accuracy} examines whether
differences in item parameter recovery and offline agreement persist at matched
test accuracy, and Section~\ref{sec:results_cost} evaluates whether the gains
require more trainable parameters.

\subsection{Item Order and Item Parameter Estimation}
\label{sec:results_trajectories}

The derivatives in Section~\ref{sec:item_estimation} show that discrimination
and option slope updates multiply the prediction residual by an ability estimate
or an ability-to-location distance, whereas difficulty, threshold, and intercept
updates do not contain those terms outside the residual. We therefore evaluate
recovery of the two parameter groups separately and measure the item-specific
ability offsets \(\phi_q\) and gains \(s_q\) associated with the affine
transformations in Section~\ref{sec:affine_invariance}.

\begin{figure}[!htbp]
\centering
\includegraphics[width=\textwidth]{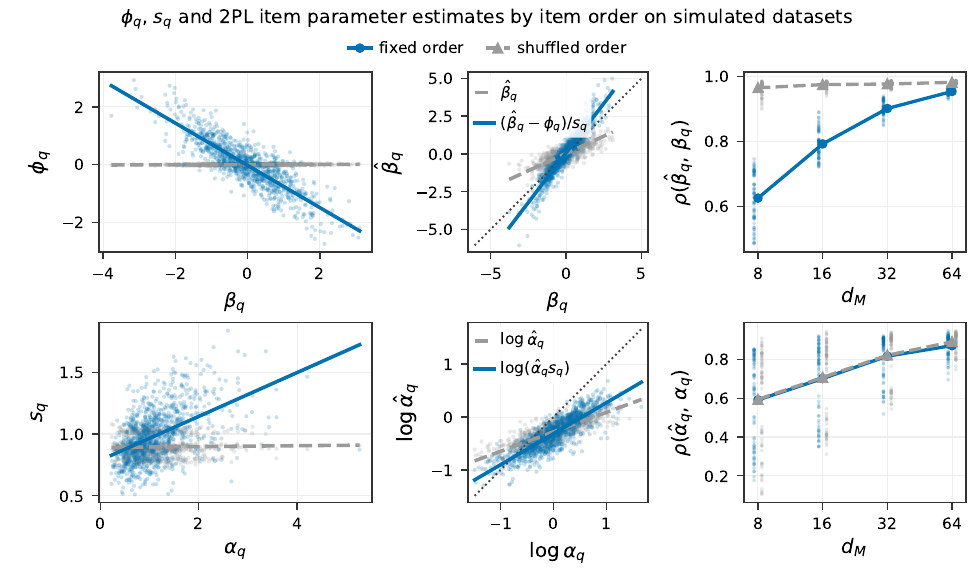}
\Description{Six panels compare fixed and shuffled item order for the 2PL.
Under fixed order, ability offsets decrease with generating difficulty and
ability gains increase with discrimination. Affine correction moves estimated
parameters toward the identity line without exact recovery. With shuffled
order, difficulty recovery is high across widths, while discrimination
recovery increases as the parameter key widens.}
\caption{Item order and 2PL parameter estimation. The left column plots
\(\phi_q\) and \(s_q\) against \(\beta_q\) and \(\alpha_q\) at
\(\mathrm{SK}(8,8)\). The middle column compares the uncorrected and
affine-corrected estimates. The right column varies \(\dM\) while \(\dS=8\).
Blue denotes fixed order and gray denotes shuffled order. Corresponding plots
for the GPCM and NRM appear in Appendix~\ref{app:order-comparison}.}
\label{fig:channels-2pl}
\end{figure}

Under the fixed-order condition, one item permutation is tiled across learners
so that each item appears at three of the 60 sequence positions and has a fixed
successor. Because the residual in Eq.~\eqref{eq:state_residual} removes only
main effects of learner and sequence position, it cannot remove an interaction
between an item and the specific positions that item occupies. The shuffled
order condition tests whether item-specific offsets and gains persist when each
learner receives an independently sampled and ordered set of 60 items.
Figure~\ref{fig:channels-2pl} illustrates these comparisons for the 2PL, while
Table~\ref{tab:traj} summarizes the results across the 2PL, GPCM, and NRM (with
the corresponding six-panel GPCM and NRM plots reported in
Figures~\ref{fig:channels-gpcm} and~\ref{fig:channels-nrm} in
Appendix~\ref{app:order-comparison}).

In the 2PL (Figure~\ref{fig:channels-2pl}), fixed item order induces a strong
negative association between the ability offset \(\phi_q\) and generating
difficulty (\(\rho=-0.84\)) and a positive association between the ability gain
\(s_q\) and discrimination (\(\rho=0.50\)). Both associations collapse toward
zero under shuffled order. Applying the affine inverse transformation in
Eq.~\eqref{eq:ordered_correction} to the run-averaged fixed-order estimates
moves the least-squares slope for difficulty from 0.46 to 1.31 and the slope for
log discrimination from 0.37 to 0.59. With \(\hat s_q\) estimated by
finite-sample regression with residual noise \(\varepsilon_{i,t}\) in
Eq.~\eqref{eq:affine_residual}, dividing \((\hat\beta_q-\phi_q)\) by \(\hat s_q\)
gives a linear slope above one, but restores rank
recovery \(\rho_{\mathrm{corr}}\) from 0.626 to 0.934 (Table~\ref{tab:traj}).
Most of the fixed-order estimation error thus tracks the
item-specific offset and gain of the ability estimates supplied to the decoder. In the right column of Figure~\ref{fig:channels-2pl}, where item
order is shuffled and \(\dM\) varies at \(\dS=8\), difficulty recovery is
already 0.965 at \(\dM=8\) and changes by at most 0.031 through \(\dM=64\),
whereas discrimination recovery rises from 0.594 to 0.892. Thus, breaking the
fixed successor relation removes the order-induced offset association but
preserves the sharp sensitivity of discrimination recovery to parameter key
width \(\dM\).

The same two patterns hold for ordered polytomous responses under the GPCM and
nominal responses under the NRM (Appendix~\ref{app:order-comparison},
Figures~\ref{fig:channels-gpcm} and~\ref{fig:channels-nrm}). For the GPCM under
fixed order, \(\rho(\phi_q,\bar\beta_q)=-0.82\) and
\(\rho(s_q,\alpha_q)=0.62\), and the affine correction moves the fitted slope
from 0.42 to 1.04 for the mean step threshold and from 0.48 to 0.73 for log
discrimination. Under shuffled order, mean threshold recovery starts at 0.978 at
\(\dM=8\) while discrimination recovery rises from 0.716 to 0.948 across
\(\dM\). For the NRM under fixed order, \(\phi_q\) correlates \(-0.82\) with the
correct-option location \(-c_{q,k_q^\star}/a_{q,k_q^\star}\) and \(s_q\)
correlates \(0.76\) with the keyed option slope contrast \(a_q^{\mathrm{key}}\),
with Eq.~\eqref{eq:nrm_correction} raising the fitted slope from 0.47 to 0.84
for the correct-option intercept and from 0.02 to 0.43 for
\(a_q^{\mathrm{key}}\). Under shuffled order, correct-option intercept recovery
is 0.946 at \(\dM=8\), whereas recovery of \(a_q^{\mathrm{key}}\) climbs from
0.349 to 0.747 as \(\dM\) increases to 64. The recovery gap between location
parameters and discrimination or option slopes, along with its persistence under shuffled order at narrow
\(\dM\), therefore extends across binary, ordinal, and nominal response models.

\begin{table}[!htbp]
\centering
\caption{Split-half agreement of the ability offset $\phi_q$ and recovery of location
parameters before and after the affine correction.
$\rho_{\mathrm{half}}$ compares $\phi_q$ across two disjoint halves of the test
split. $\rho_{\mathrm{unc}}$ and $\rho_{\mathrm{corr}}$ give uncorrected
and corrected recovery. Each entry averages 75 training runs.}
\label{tab:traj}
\small
\setlength{\tabcolsep}{4.2pt}
\begin{tabular}{@{}lccccccccc@{}}
\toprule
 & \multicolumn{3}{c}{2PL} & \multicolumn{3}{c}{GPCM}
 & \multicolumn{3}{c}{NRM} \\
\cmidrule(lr){2-4}\cmidrule(lr){5-7}\cmidrule(lr){8-10}
Width choice & $\rho_{\mathrm{half}}$ & $\rho_{\mathrm{unc}}$
     & $\rho_{\mathrm{corr}}$
     & $\rho_{\mathrm{half}}$ & $\rho_{\mathrm{unc}}$
     & $\rho_{\mathrm{corr}}$
     & $\rho_{\mathrm{half}}$ & $\rho_{\mathrm{unc}}$
     & $\rho_{\mathrm{corr}}$ \\
\midrule
\multicolumn{10}{@{}l}{\emph{Fixed item order}}\\
SK $\dM{=}8$, $\dS{=}8$            & 0.961 & 0.626 & 0.934 & 0.957 & 0.617 & 0.931 & 0.955 & 0.426 & 0.919 \\
SK $\dM{=}8$, $\dS{=}64$           & 0.977 & 0.385 & 0.916 & 0.969 & 0.341 & 0.909 & 0.960 & 0.252 & 0.915 \\
SK $\dM{=}64$, $\dS{=}8$           & 0.831 & 0.954 & 0.973 & 0.798 & 0.951 & 0.977 & 0.848 & 0.933 & 0.969 \\
SK $\dM{=}64$, $\dS{=}64$          & 0.948 & 0.848 & 0.962 & 0.920 & 0.860 & 0.963 & 0.933 & 0.838 & 0.963 \\
SH $w{=}8$                         & 0.953 & 0.660 & 0.939 & 0.952 & 0.619 & 0.934 & 0.935 & 0.623 & 0.928 \\
SH $w{=}64$                        & 0.944 & 0.830 & 0.960 & 0.919 & 0.858 & 0.961 & 0.909 & 0.865 & 0.961 \\
\addlinespace[2pt]
\multicolumn{10}{@{}l}{\emph{Shuffled item order}}\\
SK $\dM{=}8$, $\dS{=}8$            & 0.066 & 0.965 & 0.964 & 0.062 & 0.978 & 0.977 & 0.062 & 0.946 & 0.945 \\
SH $w{=}8$                         & 0.041 & 0.968 & 0.966 & 0.044 & 0.980 & 0.978 & 0.060 & 0.961 & 0.960 \\
SH $w{=}64$                        & 0.049 & 0.979 & 0.977 & 0.039 & 0.984 & 0.983 & 0.070 & 0.973 & 0.972 \\
\bottomrule
\end{tabular}
\end{table}

Table~\ref{tab:traj} quantifies the repeatability of the ability offsets
\(\phi_q\) across two disjoint halves of the test split alongside location
recovery before and after affine correction. Under fixed order, split-half
agreement \(\rho_{\mathrm{half}}\) for \(\phi_q\) ranges from 0.798 to 0.977
even when uncorrected location recovery \(\rho_{\mathrm{unc}}\) drops as low as
0.252, and applying the affine correction restores location recovery to
0.909--0.977 across all width choices and response models. Under shuffled order,
\(\rho_{\mathrm{half}}\) falls to 0.039--0.070 and the affine correction alters
location recovery by at most 0.002. Moreover, in out-of-sample checks on test
learners not used to fit the transformation, item-specific affine fits improve
upon a single common affine fit under fixed order but not under shuffled order
(Appendix~\ref{app:affine-checks}). Fixed item order thus produces a repeatable,
item-specific distortion in \(\hat\theta^-_{i,t}\) that is absorbed into the
estimated item parameters.

\begin{figure}[!htbp]
\centering
\includegraphics[width=\textwidth]{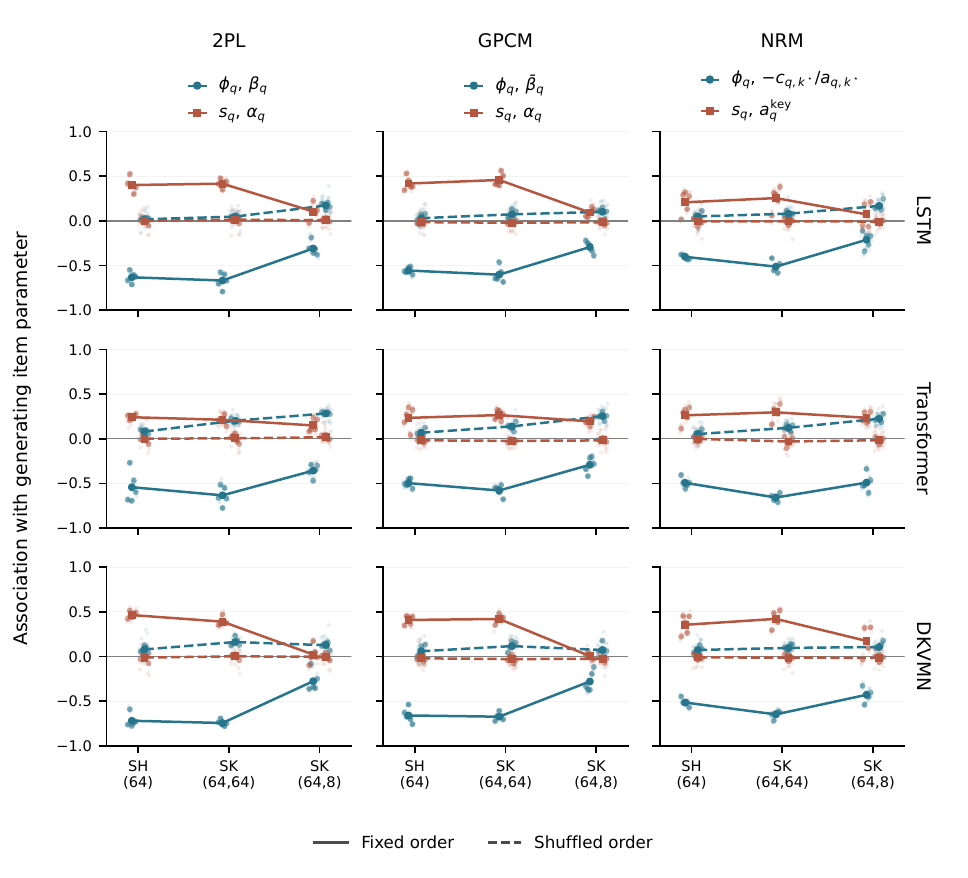}
\Description{Nine panels compare SH(64), SK(64,64), and SK(64,8) for three
encoders and response models. Solid fixed-order gain correlations move toward
the near-zero dashed shuffled-order reference under SK(64,8). Offset
correlations shift upward in both orders. Separating tables at width 64 alone
does not show the same consistent reduction in fixed-order associations.}
\caption{Ability offsets and gains under \(\mathrm{SH}(64)\),
\(\mathrm{SK}(64,64)\), and \(\mathrm{SK}(64,8)\). Colors identify the Spearman
correlation plotted. Solid lines denote fixed order and dashed lines shuffled
order. Each point on a line averages five datasets and five folds, where faint points
show individual runs and darker points show dataset means. The NRM location
comparison uses generating \(a_{q,k^\star}>0.2\).}
\label{fig:order-associations}
\end{figure}

\paragraph{Table sharing and sequence width.}
In Table~\ref{tab:traj}, under fixed item order, uncorrected location
recovery is already 0.933--0.954 for \(\mathrm{SK}(64,8)\), compared with
0.830--0.865 for \(\mathrm{SH}(64)\) and 0.838--0.860 for \(\mathrm{SK}(64,64)\).
To test systematically whether this improvement comes from separating the item
tables or from narrowing the sequence input width \(\dS\),
Figure~\ref{fig:order-associations} compares the ability--parameter associations
across \(\mathrm{SH}(64)\), \(\mathrm{SK}(64,64)\), and \(\mathrm{SK}(64,8)\)
for all nine encoder and response-model pairs. Under fixed order,
\(\mathrm{SH}(64)\) exhibits negative offset associations and positive gain
associations (ranging from 0.208 to 0.465 across the nine pairs, versus
\(-0.019\) to \(0.001\) after shuffling).

As shown in Figure~\ref{fig:order-associations}, \(\mathrm{SK}(64,8)\) weakens
the fixed-order associations relative to \(\mathrm{SH}(64)\) across encoders and
response models, pulling the fixed-order gain correlations toward the near-zero
shuffled reference and shifting the offset correlations toward zero as well. By
contrast, \(\mathrm{SK}(64,64)\) does not produce a consistent reduction.
Correspondingly, cross-order item-rank agreement \(\rho_\xi^{\mathrm{ord}}\)
between fixed and shuffled administrations improves under \(\mathrm{SK}(64,8)\)
relative to \(\mathrm{SH}(64)\) for both parameter groups in all nine encoder
and response-model pairs, rising by 0.012 to 0.230 for location parameters and by
0.008 to 0.157 for discrimination and \(a_q^{\mathrm{key}}\) (with the largest
gains in DKVMN), whereas \(\mathrm{SK}(64,64)\) shows no consistent improvement
(Appendix~\ref{app:order-comparison}, Figure~\ref{fig:order-dependence}). A wide
parameter key paired with a thin sequence representation therefore reduces
fixed-order sensitivity much more effectively than table separation at fixed
width.

The supplementary DKVMN study (Appendix~\ref{app:dkvmn-readout},
Figure~\ref{fig:dkvmn-summary}) examines a complementary source of item
dependence when the current item key enters the ability readout directly. Even
under random item order, concatenating the current item key into the ability
readout induces a negative correlation between per-item mean ability and
generating difficulty (\(-0.70\) in \(\mathrm{SH}(64)\) and \(-0.47\) in
\(\mathrm{SK}(64,8)\)), whereas removing that direct item input brings both correlations
near zero (\(0.04\) and \(0.11\)) and raises difficulty recovery from 0.856 to
0.977 in \(\mathrm{SH}(64)\) and from 0.945 to 0.983 in \(\mathrm{SK}(64,8)\).
Item-dependent variation in \(\hat\theta^-_{i,t}\) can thus enter either through
predictable interaction history or through direct current-item input to the
ability readout.

\subsection{Item Representation Width}
\label{sec:results_width}

\begin{figure}[!htbp]
\centering
\Description{Nine line plots show recovery relative to SH(8), with parameter
key width on the horizontal axis and one blue curve per sequence width.
Recovery rises with parameter key width in every panel. At parameter key
width 64, thinner sequence representations give higher recovery. Gray dashed
curves show the shared-table width comparison.}
\includegraphics[width=\textwidth]{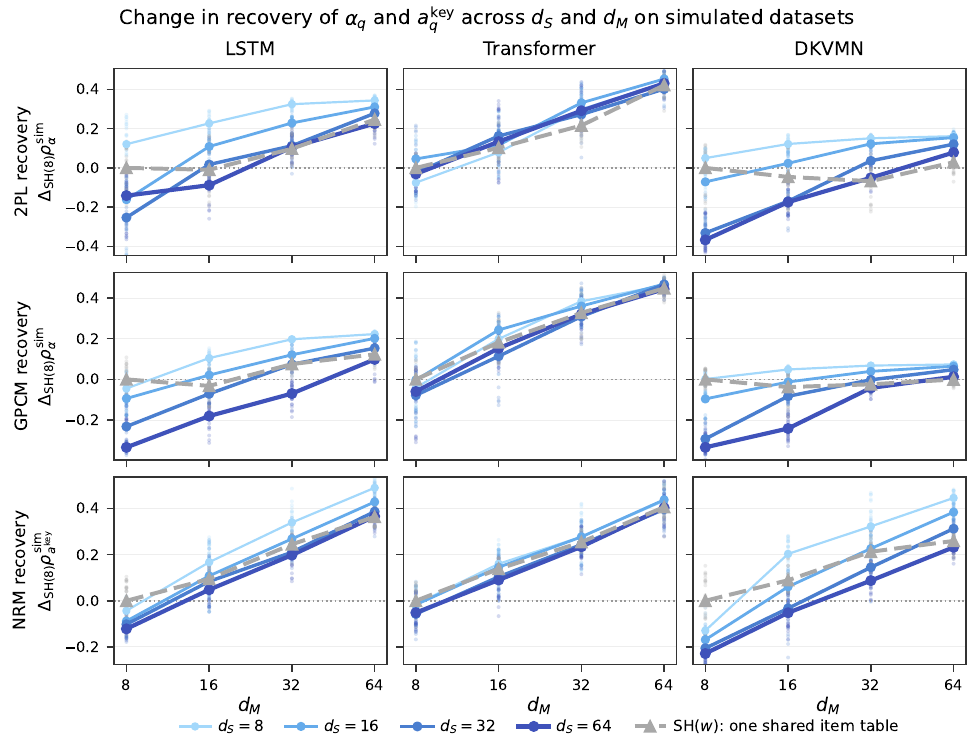}
\caption{Simulated item parameter recovery relative to SH(8). The 2PL and
GPCM rows report $\Delta_{\mathrm{SH}(8)}\rho_\alpha^{\mathrm{sim}}$, and the
NRM row reports $\Delta_{\mathrm{SH}(8)}\rho_{a^{\mathrm{key}}}^{\mathrm{sim}}$.
Blue curves hold $\dS$ fixed while $\dM$ varies, and the gray dashed curve varies
the shared width $w$ in SH. Curves join means for each width choice, and faint points show the
25 training runs. Twenty of 4{,}500 runs fall outside the displayed row
ranges. Absolute recovery is shown in Figure~\ref{fig:app-heatmap-sim} and
Appendix~\ref{app:width-study}.}
\label{fig:width-sim-rel}
\end{figure}

\begin{figure}[!htbp]
\centering
\includegraphics[width=\textwidth]{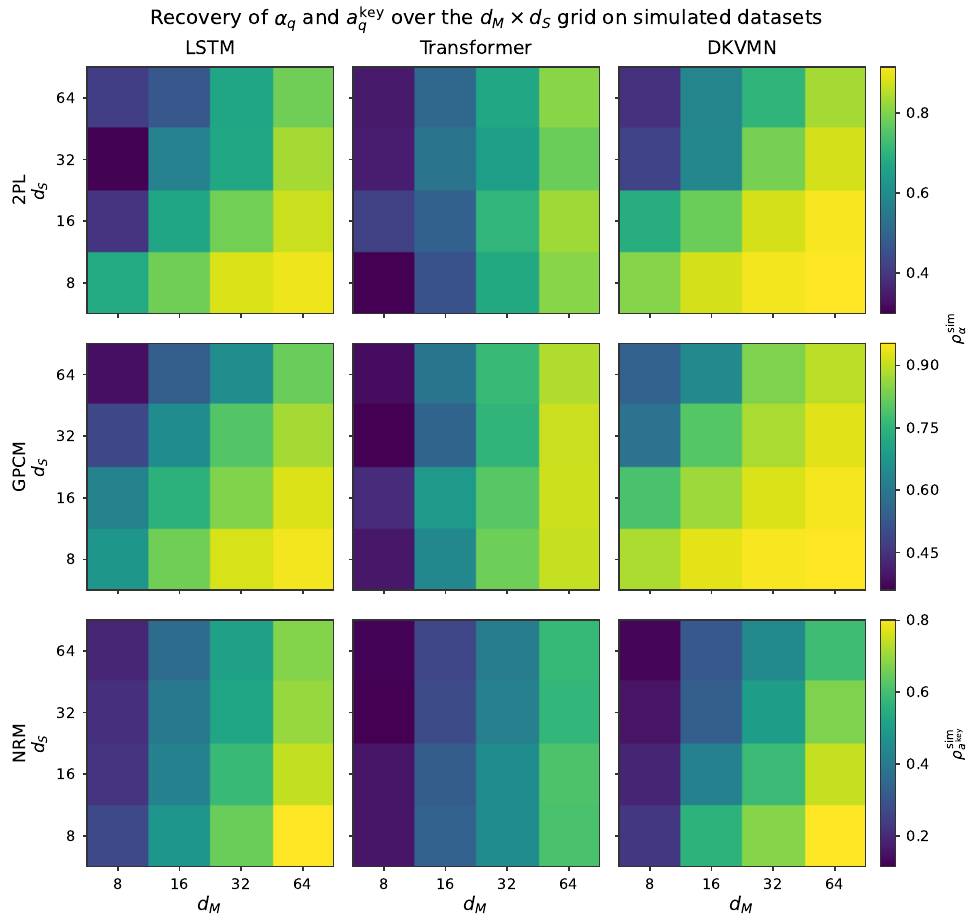}
\Description{Nine heatmaps cross four parameter key widths horizontally with
four sequence widths vertically. Brighter tiles indicate stronger simulated
recovery. Recovery increases toward wider keys, with thinner sequence
representations generally producing the brightest tiles at key width 64.
Rows correspond to response models and columns to encoders.}
\caption{Absolute recovery of \(\alpha_q\) and \(a_q^{\mathrm{key}}\) over the
complete \(\mathrm{SK}(\dM,\dS)\) width grid on simulated data. Each tile
averages 25 training runs. Color limits are shared across encoders within each
response model.}
\label{fig:app-heatmap-sim}
\end{figure}

\begin{figure}[!htbp]
\centering
\Description{Nine line plots show offline-reference agreement relative to
SH(8). Blue curves vary parameter key width at fixed sequence widths and gray
dashed curves vary shared-table width. A wider key improves the thin-sequence
endpoint comparisons, while the effect of sequence width differs across
encoders and datasets. The DKVMN EdNet 2PL curves are nearly flat.}
\includegraphics[width=\textwidth]{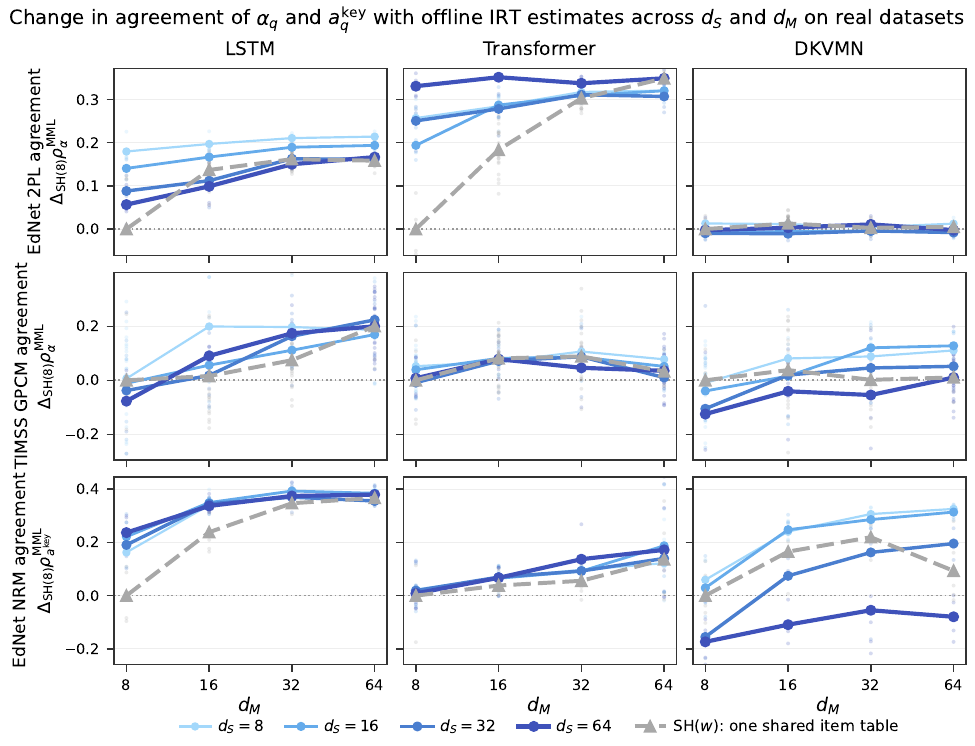}
\caption{Agreement with the offline IRT reference relative to SH(8). The 2PL
and GPCM rows report $\Delta_{\mathrm{SH}(8)}\rho_\alpha^{\mathrm{MML}}$, and
the NRM row reports
$\Delta_{\mathrm{SH}(8)}\rho_{a^{\mathrm{key}}}^{\mathrm{MML}}$. Blue curves
hold $\dS$ fixed while $\dM$ varies, and the gray dashed curve varies $w$ in SH.
Curves join means for each width choice, and faint points show individual
training runs. Eleven of 1{,}260 runs fall outside the displayed row ranges.
Absolute agreement is reported in Appendix~\ref{app:width-study}.}
\label{fig:width-real-rel}
\end{figure}

Because the order comparison favors \(\mathrm{SK}(64,8)\) over both
\(\mathrm{SH}(64)\) and \(\mathrm{SK}(64,64)\), we next vary the parameter key
width \(\dM\) and sequence input width \(\dS\) independently over the complete
\(4\times4\) grid \(\{8,16,32,64\}^2\) to separate their effects on parameter
estimation. Figures~\ref{fig:width-sim-rel} and~\ref{fig:width-real-rel} report
changes relative to \(\mathrm{SH}(8)\), \(\Delta_{\mathrm{SH}(8)}\rho_\xi\), for
each dataset, encoder, and response model, while
Figure~\ref{fig:app-heatmap-sim} displays absolute simulated recovery across the
two-dimensional \((\dM,\dS)\) grid (additional absolute curves and real-data
heatmaps appear in Appendix~\ref{app:width-study}).

On simulated data (Figures~\ref{fig:width-sim-rel}
and~\ref{fig:app-heatmap-sim}), \(\dM\) and \(\dS\) act in opposite directions.
Increasing the parameter key width \(\dM\) from 8 to 64 raises simulated
recovery for every fixed value of \(\dS\), with all 36 paired changes meeting
the predeclared decision rule in Eq.~\eqref{eq:decision_rule}. Conversely, at
\(\dM=64\), increasing the sequence input width \(\dS\) from 8 to 64 lowers
recovery in all nine encoder and response-model pairs (with four pairs exceeding
the predeclared threshold), so the brightest tile in every panel of
Figure~\ref{fig:app-heatmap-sim} lies at or near the wide-key, thin-sequence
corner \((\dM=64,\dS=8)\).

\begin{table}[!htbp]
\centering
\caption{Simulated test accuracy and parameter recovery for SK(64,8).
Differences from SH(8), SH(64), and SK(8,64) are paired and average 25 training
runs. Cost is the trainable parameter ratio to SH(64). Bootstrap intervals are
reported in Appendix~\ref{app:complete-comparisons}.}
\label{tab:sim-gaps}
\scriptsize
\setlength{\tabcolsep}{2.0pt}
\begin{tabular}{@{}ll rrrr rrrr r@{}}
\toprule
 & & \multicolumn{4}{c}{Test accuracy}
 & \multicolumn{4}{c}{Recovery, $\alpha_q$ or $a_q^{\mathrm{key}}$} & Cost \\
\cmidrule(lr){3-6}\cmidrule(lr){7-10}\cmidrule(l){11-11}
Model & Encoder & SK(64,8) & vs SH(8) & vs SH(64) & vs SK(8,64)
& SK(64,8) & vs SH(8) & vs SH(64) & vs SK(8,64) & Ratio \\
\midrule
2PL & LSTM        & .714 & +.003 & +.001 & +.004 & .898 & +.344 & +.098 & +.485 & .62 \\
2PL & Transformer & .713 & +.004 & +.000 & +.001 & .806 & +.433 & +.011 & +.464 & .96 \\
2PL & DKVMN       & .715 & -.000 & +.002 & +.002 & .914 & +.162 & +.133 & +.529 & .68 \\
\midrule
GPCM & LSTM        & .502 & +.015 & +.008 & +.028 & .941 & +.223 & +.099 & +.557 & .62 \\
GPCM & Transformer & .492 & +.033 & +.001 & +.028 & .900 & +.463 & +.015 & +.522 & .96 \\
GPCM & DKVMN       & .502 & +.006 & +.003 & +.013 & .952 & +.073 & +.074 & +.407 & .68 \\
\midrule
NRM & LSTM        & .498 & +.016 & +.002 & +.030 & .798 & +.490 & +.126 & +.611 & .62 \\
NRM & Transformer & .489 & +.040 & +.003 & +.043 & .604 & +.433 & +.027 & +.485 & .96 \\
NRM & DKVMN       & .499 & +.005 & +.005 & +.017 & .800 & +.446 & +.186 & +.674 & .68 \\
\bottomrule
\end{tabular}
\end{table}

Table~\ref{tab:sim-gaps} summarizes the matched simulated comparisons for
\(\mathrm{SK}(64,8)\) against \(\mathrm{SH}(8)\), \(\mathrm{SH}(64)\), and
\(\mathrm{SK}(8,64)\). Relative to \(\mathrm{SH}(8)\), \(\mathrm{SK}(64,8)\)
raises recovery of \(\alpha_q\) or \(a_q^{\mathrm{key}}\) by 0.073 to 0.490
across all nine pairs (meeting the predeclared rule in every pair). Relative to
\(\mathrm{SH}(64)\), it raises recovery by 0.011 to 0.186 across all nine pairs
(mean gain 0.131 for DKVMN, 0.108 for LSTM, and 0.018 for Transformer).
Relative to \(\mathrm{SK}(8,64)\), which exchanges \(\dM\) and \(\dS\) and stores
the exact same number of item table entries, \(\mathrm{SK}(64,8)\) raises
recovery by 0.407 to 0.674. All of these gains occur while test accuracy changes
by no more than 0.008 against \(\mathrm{SH}(64)\).

To test whether the recovery loss at larger \(\dS\) is driven by backward
gradient interference from later responses (Eq.~\eqref{eq:sh_gradient}), we
compare the blocked and unblocked later-response gradient conditions described
in Section~\ref{sec:experimental_setup}. Blocking later prediction losses from
updating the parameter key through the sequence encoder improves recovery of
\(\alpha_q\) or \(a_q^{\mathrm{key}}\) in all nine encoder and response-model
pairs (with seven meeting the predeclared rule under the update-before-read
implementation, whereas under the original read-before-update DKVMN ordering the
gains of \(+0.028\) to \(+0.050\) do not meet both parts of the rule). However,
recovery still decreases as \(\dS\) increases from 8 to 64 in all nine
pairs even when later-response gradients are blocked, and none of the reductions
in the width-related drop meets the predeclared rule. Thus, while backward
gradient interference can reduce recovery, blocking it does not eliminate the
penalty of a wide \(\dS\).

On the TIMSS and EdNet real-data benchmarks (Figure~\ref{fig:width-real-rel} and
Table~\ref{tab:real-gaps}), increasing \(\dM\) from 8 to 64 at \(\dS=8\)
raises offline IRT
agreement \(\rho_\xi^{\mathrm{MML}}\) in all nine pairs, as does increasing the
shared width \(w\) from 8 to 64 in \(\mathrm{SH}(w)\). In the matched
comparisons (Table~\ref{tab:real-gaps}), \(\mathrm{SK}(64,8)\) exceeds
\(\mathrm{SK}(8,64)\) in eight of the nine pairs and exceeds \(\mathrm{SH}(64)\)
in seven of the nine pairs. The two reversals against \(\mathrm{SH}(64)\) occur
for the Transformer on EdNet 2PL (\(\Delta\rho^{\mathrm{MML}}=-0.031\)) and
EdNet NRM (\(\Delta\rho^{\mathrm{MML}}=-0.014\)). Thus, while widening \(\dM\)
consistently improves agreement with the offline IRT reference, the effect of
narrowing \(\dS\) on real data depends on the dataset and encoder.

\begin{table}[!htbp]
\centering
\small
\caption{SK(64,8) test accuracy and agreement with the offline IRT reference, with paired differences from SH(64). LSTM and Transformer use 25 pairs for 2PL and GPCM and five for NRM. DKVMN uses five pairs per study. Cost is the trainable parameter ratio to SH(64).}
\label{tab:real-gaps}
\begin{tabular}{llrrrrr}
\toprule
Study & Encoder & $\mathrm{Acc}_{\rm SK}$ & $\Delta\mathrm{Acc}$ & $\rho_{\rm SK}^{\rm MML}$ & $\Delta\rho^{\rm MML}$ & Cost \\
\midrule
EdNet 2PL & LSTM & 0.641 & +0.021 & 0.846 & +0.043 & 0.46 \\
EdNet 2PL & Transformer & 0.617 & -0.018 & 0.827 & -0.031 & 0.77 \\
EdNet 2PL & DKVMN & 0.653 & +0.003 & 0.856 & +0.006 & 0.48 \\
TIMSS GPCM & LSTM & 0.579 & -0.002 & 0.325 & +0.027 & 0.35 \\
TIMSS GPCM & Transformer & 0.580 & -0.001 & 0.518 & +0.022 & 0.89 \\
TIMSS GPCM & DKVMN & 0.584 & +0.006 & 0.545 & +0.090 & 0.37 \\
EdNet NRM & LSTM & 0.590 & +0.010 & 0.681 & +0.020 & 1.07 \\
EdNet NRM & Transformer & 0.590 & -0.001 & 0.213 & -0.014 & 1.09 \\
EdNet NRM & DKVMN & 0.596 & +0.002 & 0.686 & +0.233 & 1.08 \\
\bottomrule
\end{tabular}
\end{table}

\subsection{Prediction Accuracy and Item Parameter Estimation}
\label{sec:results_accuracy}

\begin{figure}[!htbp]
\centering
\includegraphics[width=\textwidth]{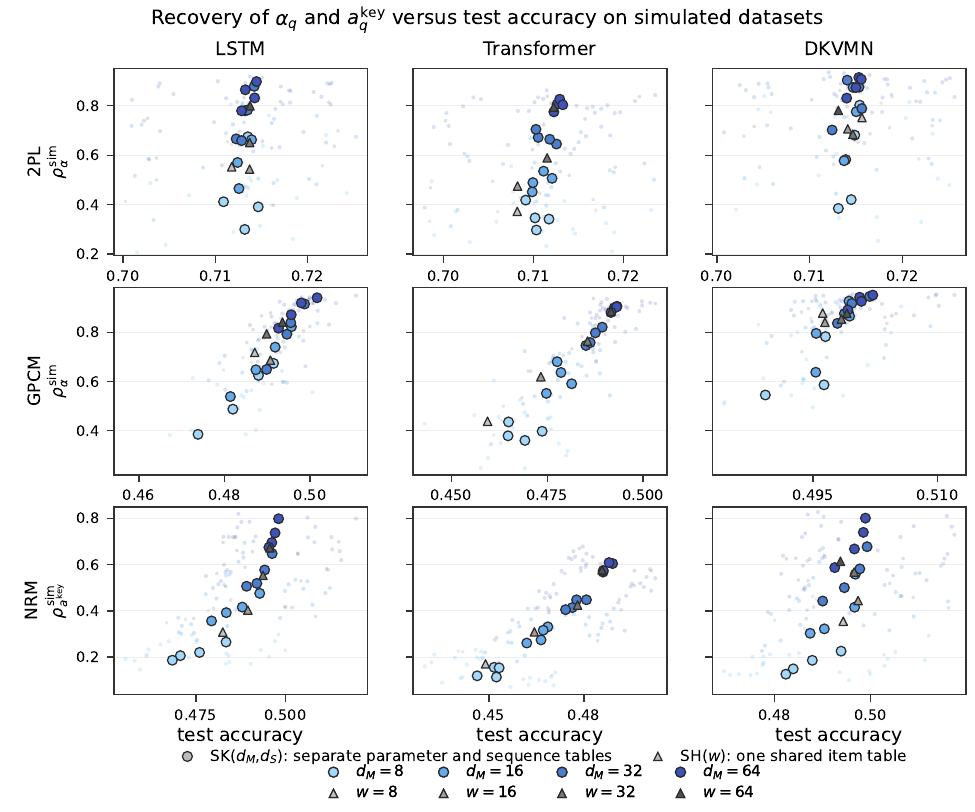}
\Description{Nine scatterplots place simulated discrimination or keyed
contrast recovery against test accuracy. Circles represent SK and triangles
SH, with item width encoded by color. Recovery varies substantially among
models with similar test accuracy. Wider parameter keys generally have higher
recovery, and faint points show variation across generated datasets.}
\caption{Test accuracy and recovery of 2PL and GPCM discrimination
\(\alpha_q\) and the NRM keyed option slope contrast
\(a_q^{\mathrm{key}}\) on simulated
data. Circles denote SK and triangles denote SH. Color indicates \(\dM\) for SK
and \(w\) for SH; large markers show width-choice means and faint markers show
means for the five generated datasets.}
\label{fig:accuracy-sim}
\end{figure}

\begin{figure}[!htbp]
\centering
\includegraphics[width=\textwidth]{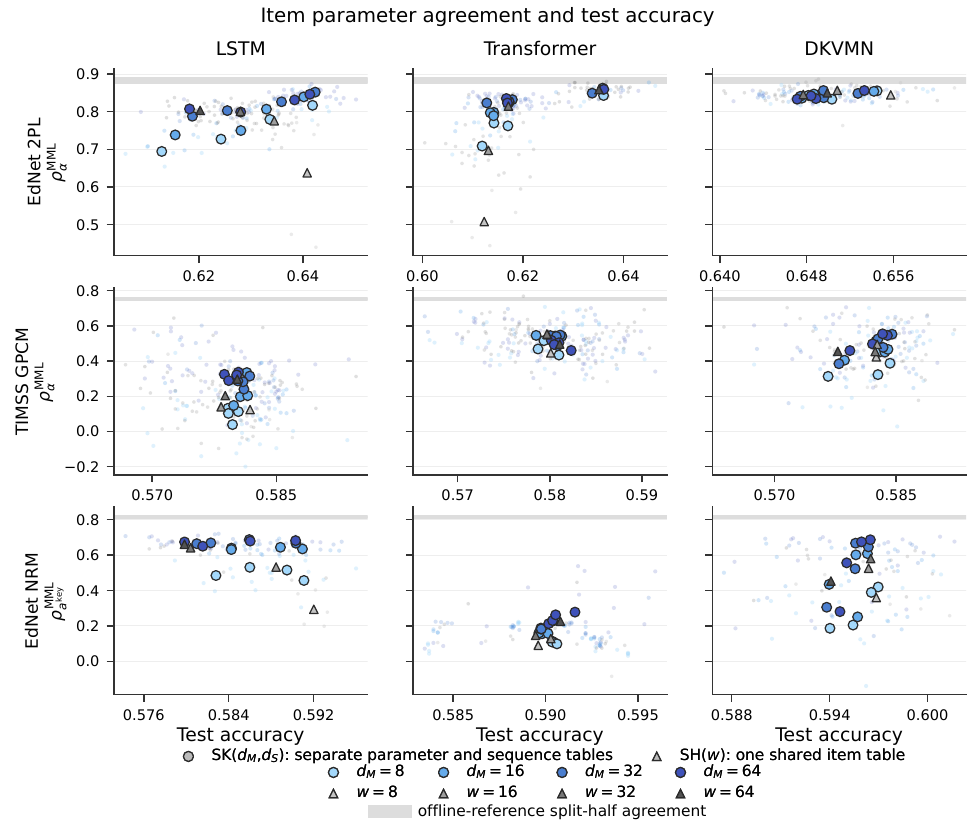}
\Description{Nine scatterplots compare real-data agreement with an offline
IRT reference against test accuracy. SK circles and SH triangles can have
similar accuracy but different item parameter agreement. The patterns differ
across encoders and datasets. Faint run points show the spread, and a gray
horizontal reference gives agreement between two offline estimates.}
\caption{Test accuracy and agreement of \(\alpha_q\) and
\(a_q^{\mathrm{key}}\) with the offline IRT reference on real data. Markers
follow Figure~\ref{fig:accuracy-sim}, with faint markers showing individual training
runs and the gray line marking split-half agreement between two offline estimates from
separate response halves.}
\label{fig:accuracy-real}
\end{figure}

To test whether stronger item parameter estimation simply reflects better
response prediction, we next compare parameter recovery and offline agreement against
next-response test accuracy across all 20 width choices
(Figures~\ref{fig:accuracy-sim} and~\ref{fig:accuracy-real}).

In simulation (Figure~\ref{fig:accuracy-sim}), models with nearly identical test
accuracy assign widely different rankings to discrimination \(\alpha_q\) and the
NRM keyed option slope contrast \(a_q^{\mathrm{key}}\). In the 2PL panels, for
example, test accuracy across all 20 width choices clusters within a narrow band
around 0.71, while \(\rho_\alpha^{\mathrm{sim}}\) spans from roughly 0.30 at
\(\dM=8\) to above 0.90 at \(\dM=64\). Across all nine encoder and
response-model pairs, moving from \(\mathrm{SH}(8)\) to \(\mathrm{SK}(64,8)\)
raises recovery by 0.073 to 0.490 while test accuracy changes by less than 0.05
in every pair (and by at most 0.004 in the 2PL). High next-response accuracy
therefore does not guarantee accurate recovery of slope or discrimination
parameters.

On TIMSS and EdNet (Figure~\ref{fig:accuracy-real}), comparing
\(\mathrm{SK}(64,8)\) with \(\mathrm{SH}(8)\) changes test accuracy by only
\(-0.002\) to \(+0.002\), yet offline agreement \(\rho_\xi^{\mathrm{MML}}\)
increases in all nine pairs by 0.012 to 0.387. At the same time,
Figure~\ref{fig:accuracy-real} and Table~\ref{tab:real-gaps} clarify the two
real-data reversals against \(\mathrm{SH}(64)\) noted in
Section~\ref{sec:results_width}. For the EdNet 2PL Transformer, where
\(\mathrm{SK}(64,8)\) lowers offline agreement by 0.031 relative to
\(\mathrm{SH}(64)\), it also lowers test accuracy by 0.018 (and increases test
log-loss by 0.068 in Appendix~\ref{app:probability-checks},
Table~\ref{tab:real-rank-prediction}), whereas for the EdNet 2PL LSTM both test
accuracy (\(+0.021\)) and log-loss (\(-0.076\)) improve alongside offline
agreement (\(+0.043\)). Thus, the EdNet 2PL Transformer reversal accompanies a
simultaneous change in predictive performance when \(\dS\) is compressed from 64
to 8, rather than representing a drop in agreement at fixed prediction quality.

\subsection{Trainable Parameters and Matched Comparisons}
\label{sec:results_cost}

\begin{figure}[!htbp]
\centering
\Description{Six panels compare recovery or offline agreement gains relative
to SH(8) with trainable parameter count relative to the same baseline.
Colors distinguish encoders, circles denote SK, and triangles denote SH.
Arrows join increasing widths. In simulation, SK with a wide parameter key
and thin sequence representation improves recovery over wide SH with fewer
parameters. Real-data agreement and cost comparisons vary by dataset and encoder.}
\includegraphics[width=0.94\textwidth]{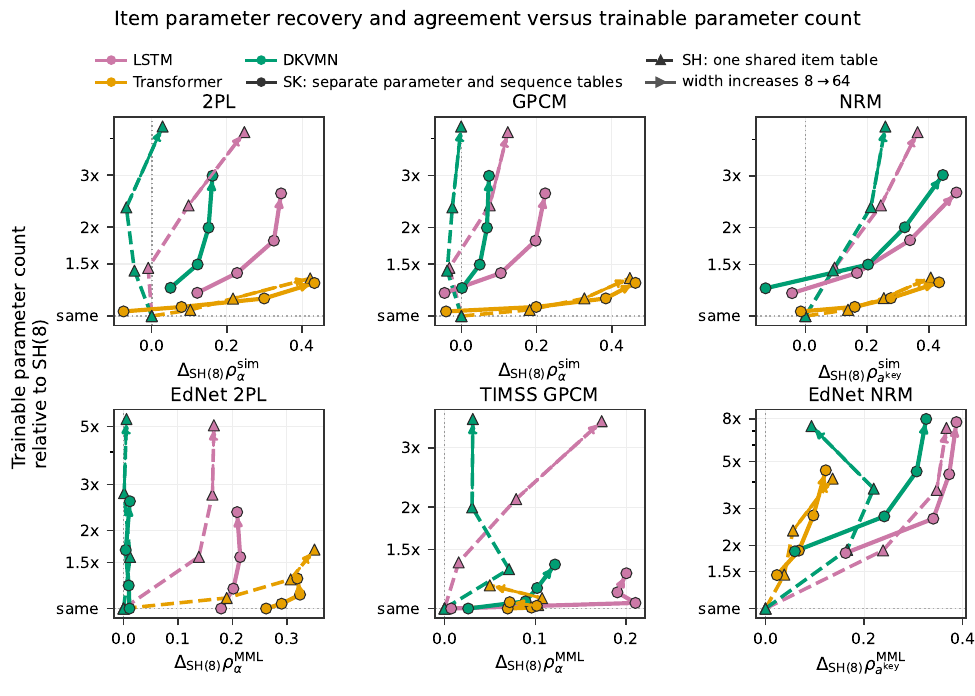}
\caption{Item parameter recovery, offline agreement, and trainable parameter count relative to
SH(8). Horizontal position is $\Delta_{\mathrm{SH}(8)}\rho_\xi$, using
simulated recovery in the top row and agreement with the offline IRT reference
in the bottom row. The vertical axis is logarithmic. Solid SK curves vary
$d_M$ with $d_S=8$, while dashed SH curves vary $w$. Arrows follow increasing
width from 8 to 64. Absolute recovery and agreement are reported in
Appendix~\ref{app:param-counts}.}
\label{fig:recovery-cost}
\end{figure}

Because varying \(\dM\) and \(\dS\) also changes total model capacity,
Figure~\ref{fig:recovery-cost} plots the recovery and offline agreement gains
\(\Delta_{\mathrm{SH}(8)}\rho_\xi\) against the ratio of trainable parameters
relative to the \(\mathrm{SH}(8)\) baseline (with absolute recovery curves and
exact parameter counts reported in Appendix~\ref{app:param-counts}). Rightward
movement indicates stronger parameter recovery or offline agreement, while
upward movement indicates a larger trainable parameter budget.

In the simulated panels (top row of Figure~\ref{fig:recovery-cost}), widening a
shared table \(\mathrm{SH}(w)\) from \(w=8\) to \(64\) (dashed curves with
triangles) drives a steep vertical increase in trainable parameters because
\(w\) simultaneously expands the sequence encoder input and the parameter
readout. By contrast, widening \(\dM\) from 8 to 64 while holding \(\dS=8\)
fixed in \(\mathrm{SK}(\dM,8)\) (solid curves with circles) traces a much
flatter cost trajectory while shifting further to the right. As reported in
Table~\ref{tab:sim-gaps}, \(\mathrm{SK}(64,8)\) uses only 62\% as many trainable
parameters as \(\mathrm{SH}(64)\) for the LSTM, 96\% for the Transformer, and
68\% for DKVMN, while achieving higher recovery in every pair.

On the real-data benchmarks (bottom row of Figure~\ref{fig:recovery-cost} and
Table~\ref{tab:real-gaps}), \(\mathrm{SK}(64,8)\) uses 35\% to 89\% as many
trainable parameters as \(\mathrm{SH}(64)\) on EdNet 2PL and TIMSS GPCM and
achieves higher offline agreement in five of those six pairs. On EdNet NRM,
where two 250-item lookup tables across four response options dominate the
parameter budget, \(\mathrm{SK}(64,8)\) uses 107\% to 109\% as many parameters
as \(\mathrm{SH}(64)\) and yields higher agreement in two of the three pairs.
Across all nine simulated comparisons and five of the six smaller-budget
real-data comparisons, the gains of \(\mathrm{SK}(64,8)\) over
\(\mathrm{SH}(64)\) are achieved with fewer trainable parameters. The
improvement therefore comes from how representation width is allocated between
\(\dM\) and \(\dS\) rather than from a larger model size.

\section{Discussion}\label{sec:discussion}

Across binary (2PL), polytomous (GPCM), and nominal (NRM) response decoders,
our results demonstrate that accurate next-response prediction and accurate
location parameter estimation often coexist with poor recovery of
discrimination and option slope parameters. This divergence reflects two
structural properties of end-to-end KT training, namely a first-order gradient
asymmetry in which discrimination and slope updates depend directly on the
spread and scaling of estimated abilities while location updates lack an ability
multiplier outside the prediction residual, and affine logit invariance, which allows
item-dependent shifts and rescaling in estimated abilities to be absorbed into
the item parameters without altering response probabilities. Separating the item
parameter key width (\(\dM\)) from the sequence representation width (\(\dS\))
in \(\mathrm{SK}(64,8)\) addresses both properties in simulation, providing
sufficient capacity for the parameter readout while limiting order-dependent
ability distortions, with fewer trainable parameters than \(\mathrm{SH}(64)\)
and no practically meaningful loss in test accuracy.

\subsection{Why Predictive Accuracy Does Not Validate Item Parameters}

Earlier studies at the intersection of KT and IRT showed that a neural KT
predictor can adopt a 1PL
response decoder \citep{yeung_deepirt_2019,tsutsumi_independent_2021} or be
written algebraically as a dynamic multidimensional IRT model
\citep{vie_kashima_2023}, while variational IRT and BKT studies observed that
predictive accuracy and parameter or skill-assignment recovery can rank models
differently \citep{wu_variational_2020,khajah_supercharging_2024}. Our findings
establish a sharper distinction \emph{within} the parameters of a single IRT
decoder, showing that even under identical training data, objective, and item-order
conditions, high recovery of difficulty, mean step thresholds, or correct-option
intercepts does not imply that the model has recovered item discrimination or
option slopes.

The response-model derivatives explain why this gap arises even when item order
is shuffled. Because difficulty, step thresholds, and option intercepts enter
the gradient without an ability multiplier outside the prediction residual, they
are anchored directly by marginal category rates. At
\(\hat\theta^-_{i,t}=\beta_q\), a 2PL response provides positive Fisher
information for difficulty (\(I_{\beta\beta}>0\)) but zero Fisher information
for discrimination (\(I_{\alpha\alpha}=0\)). Discrimination and option slope
updates, by contrast, require co-variation between the prediction residual and
the centered ability estimate, making them sensitive to both the resolution of
the parameter key \(\dM\) and any distortion in the ability distribution.

When item order is fixed, affine logit invariance introduces a second source of error.
Because joint transformations of ability (\(\hat\theta^-_{i,t}=s_q\theta^{\mathrm{ref}}_{i,t}+\phi_q\))
and item parameters leave the decoder logits unchanged, the prediction loss has
a zero-gradient null direction along which item-specific ability offsets
\(\phi_q\) and gains \(s_q\) trade off directly against estimated item locations
and slopes. In our fixed-order simulations, \(\phi_q\) is highly repeatable
across independent halves of the test split (\(\rho_{\mathrm{half}}\in[0.80,0.98]\)),
predicts out-of-sample test-learner ability estimates better than a common
affine fit, and accounts for the bulk of the uncorrected location error when
inverted. Next-response prediction accuracy is insensitive to this compensating
shift, which is why models with nearly identical test accuracy can assign
markedly different rankings to item parameters.

\subsection{Readout Capacity Versus Sequence Representation Width}

Comparing \(\mathrm{SH}(64)\), \(\mathrm{SK}(64,64)\), and \(\mathrm{SK}(64,8)\)
clarifies why separating representation widths is more effective than table
separation alone. Separating the sequence and parameter tables at width 64
(\(\mathrm{SK}(64,64)\)) prevents parameter-readout gradients from updating the
sequence table, yet it does not consistently reduce fixed-order ability--parameter
associations or improve cross-order rank agreement \(\rho_\xi^{\mathrm{ord}}\).
Conversely, blocking backward gradients from later responses in Section~\ref{sec:results_width}
improves overall recovery under update-before-read DKVMN, but does not eliminate
the decline in recovery as \(\dS\) increases from 8 to 64. Together, these two
findings indicate that the penalty of a wide \(\dS\) is not solely a
backward-pass gradient artifact. Instead, a wide sequence representation allows
the encoder's forward pass to propagate predictable item-history cues into
\(\hat\theta^-_{i,t}\). Pairing a narrow sequence bottleneck (\(\dS=8\)) with a wide
parameter key (\(\dM=64\)) in \(\mathrm{SK}(64,8)\) restricts this forward
history channel while preserving the readout capacity needed to resolve
discrimination and option slopes, outperforming \(\mathrm{SK}(8,64)\) at equal
table storage and \(\mathrm{SH}(64)\) with fewer trainable parameters.

The supplementary DKVMN ablation (Appendix~\ref{app:dkvmn-readout}) identifies a
parallel forward-pass channel when the current item key enters the ability
readout directly. Under random item order, where preceding history carries no
information about the upcoming item, passing the current item key into the DKVMN
ability readout still induces a strong negative correlation between per-item
mean ability and generating difficulty (\(-0.70\) in \(\mathrm{SH}(64)\) and
\(-0.47\) in \(\mathrm{SK}(64,8)\)). Removing that direct input reduces the
correlation to near zero and restores difficulty recovery to 0.977--0.983 in
both designs, indicating that item-specific ability shifts can arise from
current-item shortcuts into the ability estimate \(\hat\theta^-_{i,t}\), whether
they enter directly through the readout or indirectly through a wide
sequence representation.

On the TIMSS and EdNet benchmarks, widening the parameter key \(\dM\) from 8 to
64 at \(\dS=8\) improves agreement with the offline IRT reference across all
nine encoder and response-model pairs, and \(\mathrm{SK}(64,8)\) outperforms
\(\mathrm{SK}(8,64)\) in eight of nine pairs and \(\mathrm{SH}(64)\) in seven of
nine pairs. The two reversals against \(\mathrm{SH}(64)\) on the EdNet
Transformer illustrate this trade-off: on EdNet 2PL, narrowing
\(\dS\) from 64 to 8 in the Transformer lowers test accuracy by
0.018 and increases test log-loss by 0.068. Whereas a thin sequence input
(\(\dS=8\)) is sufficient in controlled unidimensional simulations, complex
real-world interaction logs can require wider sequence representations for
certain encoders to maintain predictive quality. Making \(\dM\) and \(\dS\)
independently adjustable allows practitioners to tune this balance rather than
forcing both pathways to share a single width \(w\).

\subsection{Implications for Knowledge Tracing Practice}

Interpretability claims in KT are not interchangeable, because an attention weight
describes how past interactions are aggregated, a prerequisite graph describes
relations among skills, and an IRT item parameter describes a specific
structural role in a measurement equation. Consequently, demonstrating that a KT
decoder has an IRT functional form \citep{vie_kashima_2023} or achieves competitive
test accuracy does not validate its learned weights as item parameters.

Our findings suggest three practical guidelines for KT studies that interpret
decoder parameters across binary (2PL), polytomous (GPCM), or nominal (NRM)
responses.
\begin{enumerate}
\item \textbf{Evaluate item parameters separately alongside
prediction metrics.} Because difficulty, step thresholds, and intercepts can be
well recovered even when discrimination and option slopes fail, reporting only
1PL difficulty or test AUC/accuracy leaves large errors in discrimination and
option slopes undetected.
\item \textbf{Test sensitivity to item order and ability shortcuts.} Where
item administration follows a structured curriculum or fixed sequence, checking
cross-order consistency and inspecting whether per-item mean ability estimates
correlate with item difficulty helps diagnose whether ability offsets are absorbed into the
decoder.
\item \textbf{Separate parameter readout width from sequence width and report
parameter counts.} Separating \(\dM\) from \(\dS\) allows a wide parameter key
without inflating sequence encoder capacity, and reporting trainable parameter
counts ensures that improvements in parameter recovery are distinguished from
simple increases in model size.
\end{enumerate}

\subsection{Limitations and Future Work}

Our study has three main limitations. First, exact parameter
recovery can only be verified in simulation, where responses are generated from
a stationary unidimensional IRT model matching the decoder family. Our fixed
item order tests deterministic item transitions
rather than an adaptive or mastery-based curriculum, and the shuffled condition
samples item subsets independently across learners rather than permuting an
identical response matrix. Second, on TIMSS and EdNet, offline MML estimates
provide a benchmark ordering calibrated on the same response data rather than
independent ground truth, and TIMSS responses are ordered by item identifier
because administration timestamps are unavailable. Third, all models are trained
for a fixed budget of 120 epochs, and our NRM readout constrains the keyed
option slope contrast \(a_q^{\mathrm{key}}\) to be nonnegative.

Future work should extend these identifiability and width analyses to
chronological curricula with learning, forgetting, and adaptive item selection,
where item exposure depends endogenously on past performance. Incorporating
external item-calibration anchors, such as pre-calibrated item banks or
independent pretest samples, and tracking parameter trajectories across longer
training horizons will further clarify how sequential predictors can support
reliable item parameter estimation.

\section{Conclusion}\label{sec:conclusion}

When knowledge tracing models estimate IRT item parameters end to end through
next-response prediction loss, high predictive accuracy and strong recovery of
item difficulty, step thresholds, or option intercepts do not guarantee
accurate recovery of item discrimination or nominal option slopes. First-order
response derivatives explain why discrimination and option slopes depend directly on the
spread of estimated abilities, while affine logit identities show how
item-specific ability offsets and gains induced by fixed item order are absorbed
into the item parameters without penalizing prediction loss. Separating the item
parameter key (width \(\dM\)) from the sequence item representation (width
\(\dS\)) breaks the width coupling of shared-embedding architectures. In
simulation, pairing a wide parameter key with a thin sequence representation
(\(\mathrm{SK}(64,8)\)) reduces fixed-order ability distortions, improves
cross-order rank agreement, and substantially increases slope and discrimination
recovery with fewer trainable parameters than \(\mathrm{SH}(64)\) and no
practically meaningful loss in test accuracy. On TIMSS and EdNet, widening
\(\dM\) consistently improves agreement with offline IRT calibrations, while
comparisons across \(\dS\) show that sequence width must balance order
sensitivity against the encoder's predictive capacity. Evaluating item
parameter estimates separately from response prediction and decoupling readout
width from sequence width are therefore essential when interpreting IRT
decoders in sequential learner models.

\FloatBarrier
\section*{Declaration of Generative AI Software Tools in the Writing Process}
ChatGPT and Gemini were used to suggest title candidates and for drafting assistance, revision, and language editing
in Sections~1--6 and the appendix, and assisted with figure and table preparation.
The authors reviewed and edited the resulting material and take full
responsibility for the content of the publication.

\bibliographystyle{acmtrans}
\bibliography{references}

@inproceedings{liu_pykt_2022,
  author    = {Liu, Zitao and Liu, Qiongqiong and Chen, Jiahao and Huang, Shuyan and Tang, Jiliang and Luo, Weiqi},
  title     = {{pyKT}: A {Python} Library to Benchmark Deep Learning Based Knowledge Tracing Models},
  booktitle = {Advances in Neural Information Processing Systems (NeurIPS), Datasets and Benchmarks Track},
  volume    = {35},
  year      = {2022},
  note      = {\url{https://arxiv.org/abs/2206.11460}},
}

@article{bai_survey_2024,
  author  = {Bai, Yanhong and Zhao, Jiabao and Wei, Tingjiang and Cai, Qing and He, Liang},
  title   = {A Survey of Explainable Knowledge Tracing},
  journal = {Applied Intelligence},
  volume  = {54},
  pages   = {6483--6514},
  year    = {2024},
  doi     = {10.1007/s10489-024-05509-8},
  note      = {\url{https://doi.org/10.1007/s10489-024-05509-8}},
}

@inproceedings{piech_deep_2015,
  author    = {Piech, Chris and Bassen, Jonathan and Huang, Jonathan and Ganguli, Surya and Sahami, Mehran and Guibas, Leonidas J. and Sohl-Dickstein, Jascha},
  title     = {Deep Knowledge Tracing},
  booktitle = {Advances in Neural Information Processing Systems (NeurIPS)},
  volume    = {28},
  pages     = {505--513},
  year      = {2015},
  note      = {\url{https://arxiv.org/abs/1506.05908}},
}

@inproceedings{zhang_dkvmn_2017,
  author    = {Zhang, Jiani and Shi, Xingjian and King, Irwin and Yeung, Dit-Yan},
  title     = {Dynamic Key-Value Memory Networks for Knowledge Tracing},
  booktitle = {Proceedings of the 26th International Conference on World Wide Web (WWW)},
  pages     = {765--774},
  year      = {2017},
  doi       = {10.1145/3038912.3052580},
  note      = {\url{https://doi.org/10.1145/3038912.3052580}},
}

@inproceedings{pandey_sakt_2019,
  author    = {Pandey, Shalini and Karypis, George},
  title     = {A Self-Attentive Model for Knowledge Tracing},
  booktitle = {Proceedings of the 12th International Conference on Educational Data Mining (EDM)},
  pages     = {384--389},
  year      = {2019},
  note      = {\url{https://arxiv.org/abs/1907.06837}},
}

@inproceedings{ghosh_akt_2020,
  author    = {Ghosh, Aritra and Heffernan, Neil and Lan, Andrew S.},
  title     = {Context-Aware Attentive Knowledge Tracing},
  booktitle = {Proceedings of the 26th ACM SIGKDD Conference on Knowledge Discovery \& Data Mining (KDD)},
  pages     = {2330--2339},
  year      = {2020},
  doi       = {10.1145/3394486.3403282},
  note      = {\url{https://doi.org/10.1145/3394486.3403282}},
}

@inproceedings{liu_simplekt_2023,
  author    = {Liu, Zitao and Liu, Qiongqiong and Chen, Jiahao and Huang, Shuyan and Luo, Weiqi},
  title     = {{simpleKT}: A Simple But Tough-to-Beat Baseline for Knowledge Tracing},
  booktitle = {International Conference on Learning Representations (ICLR)},
  year      = {2023},
  note      = {\url{https://arxiv.org/abs/2302.06881}},
}

@inproceedings{yeung_deepirt_2019,
  author    = {Yeung, Chun-Kit},
  title     = {{Deep-IRT}: Make Deep Learning Based Knowledge Tracing Explainable Using Item Response Theory},
  booktitle = {Proceedings of the 12th International Conference on Educational Data Mining (EDM)},
  year      = {2019},
  note      = {\url{https://arxiv.org/abs/1904.11738}},
}

@article{khajah_supercharging_2024,
  author  = {Khajah, Mohammad M.},
  title   = {Supercharging {BKT} with Multidimensional Generalizable {IRT} and Skill Discovery},
  journal = {Journal of Educational Data Mining},
  volume  = {16},
  number  = {1},
  pages   = {233--278},
  year    = {2024},
  doi     = {10.5281/zenodo.11235602},
  note    = {\url{https://doi.org/10.5281/zenodo.11235602}},
}

@inproceedings{wu_variational_2020,
  author    = {Wu, Mike and Davis, Richard L. and Domingue, Benjamin W. and Piech, Chris and Goodman, Noah},
  title     = {Variational Item Response Theory: Fast, Accurate, and Expressive},
  booktitle = {Proceedings of the 13th International Conference on Educational Data Mining (EDM)},
  pages     = {257--268},
  year      = {2020},
  note      = {\url{https://arxiv.org/abs/2002.00276}},
}

@inproceedings{kim_variational_2023,
  author    = {Kim, Yunsung and Sankaranarayanan, Sreechan and Piech, Chris and Thille, Candace},
  title     = {Variational Temporal {IRT}: Fast, Accurate, and Explainable Inference of Dynamic Learner Proficiency},
  booktitle = {Proceedings of the 16th International Conference on Educational Data Mining (EDM)},
  year      = {2023},
  pages     = {260--268},
  doi       = {10.5281/zenodo.8115687},
  note      = {\url{https://doi.org/10.5281/zenodo.8115687}},
}

@inproceedings{curi_ml2pvae_2019,
  author    = {Curi, Mariana and Converse, Geoffrey A. and Hajewski, Jeff and Oliveira, Suely},
  title     = {Interpretable Variational Autoencoders for Cognitive Models},
  booktitle = {Proceedings of the International Joint Conference on Neural Networks (IJCNN)},
  pages     = {1--8},
  year      = {2019},
  doi       = {10.1109/IJCNN.2019.8852333},
  note      = {\url{https://doi.org/10.1109/IJCNN.2019.8852333}},
}

@article{converse_estimation_2021,
  author  = {Converse, Geoffrey and Curi, Mariana and Oliveira, Suely and Templin, Jonathan},
  title   = {Estimation of Multidimensional Item Response Theory Models with Correlated Latent Variables Using Variational Autoencoders},
  journal = {Machine Learning},
  volume  = {110},
  number  = {6},
  pages   = {1463--1480},
  year    = {2021},
  doi     = {10.1007/s10994-021-06005-7},
  note      = {\url{https://doi.org/10.1007/s10994-021-06005-7}},
}

@article{muraki_gpcm_1992,
  author  = {Muraki, Eiji},
  title   = {A Generalized Partial Credit Model: Application of an {EM} Algorithm},
  journal = {Applied Psychological Measurement},
  volume  = {16},
  number  = {2},
  pages   = {159--176},
  year    = {1992},
  doi     = {10.1177/014662169201600206},
  note      = {\url{https://doi.org/10.1177/014662169201600206}},
}

@article{bock_aitkin_1981,
  author  = {Bock, R. Darrell and Aitkin, Murray},
  title   = {Marginal Maximum Likelihood Estimation of Item Parameters: Application of an {EM} Algorithm},
  journal = {Psychometrika},
  volume  = {46},
  pages   = {443--459},
  year    = {1981},
  doi     = {10.1007/BF02293801},
  note      = {\url{https://doi.org/10.1007/BF02293801}},
}

@article{bock_1972,
  author  = {Bock, R. Darrell},
  title   = {Estimating Item Parameters and Latent Ability When Responses Are Scored in Two or More Nominal Categories},
  journal = {Psychometrika},
  volume  = {37},
  number  = {1},
  pages   = {29--51},
  year    = {1972},
  doi     = {10.1007/BF02291411},
  note      = {\url{https://doi.org/10.1007/BF02291411}},
}

@article{thissen_steinberg_1986,
  author  = {Thissen, David and Steinberg, Lynne},
  title   = {A Taxonomy of Item Response Models},
  journal = {Psychometrika},
  volume  = {51},
  pages   = {567--577},
  year    = {1986},
  doi     = {10.1007/BF02295596},
  note      = {\url{https://doi.org/10.1007/BF02295596}},
}

@book{deayala_theory_2009,
  author    = {de Ayala, R. J.},
  title     = {The Theory and Practice of Item Response Theory},
  publisher = {Guilford Press},
  year      = {2009},
  isbn      = {978-1-59385-869-8},
}

@article{chalmers_mirt_2012,
  author  = {Chalmers, R. Philip},
  title   = {{mirt}: A Multidimensional Item Response Theory Package for the {R} Environment},
  journal = {Journal of Statistical Software},
  volume  = {48},
  number  = {6},
  pages   = {1--29},
  year    = {2012},
  doi     = {10.18637/jss.v048.i06},
  note      = {\url{https://doi.org/10.18637/jss.v048.i06}},
}

@inproceedings{choi_ednet_2020,
  author    = {Choi, Youngduck and Lee, Youngnam and Shin, Dongmin and Cho, Junghyun and Park, Seoyon and Lee, Seewoo and Baek, Jineon and Bae, Chan and Kim, Byungsoo and Heo, Jaewe},
  title     = {{EdNet}: A Large-Scale Hierarchical Dataset in Education},
  booktitle = {Artificial Intelligence in Education (AIED)},
  pages     = {69--73},
  year      = {2020},
  doi       = {10.1007/978-3-030-52240-7_13},
  note      = {\url{https://doi.org/10.1007/978-3-030-52240-7_13}},
}

@book{baker_kim_2004,
  author    = {Baker, Frank B. and Kim, Seock-Ho},
  title     = {Item Response Theory: Parameter Estimation Techniques},
  edition   = {2nd},
  publisher = {Marcel Dekker},
  address   = {New York},
  year      = {2004},
}

@book{lord_1980,
  author    = {Lord, Frederic M.},
  title     = {Applications of Item Response Theory to Practical Testing Problems},
  publisher = {Lawrence Erlbaum Associates},
  address   = {Hillsdale, NJ},
  year      = {1980},
}

@inproceedings{ghosh_option_2021,
  author    = {Ghosh, Aritra and Raspat, Jay and Lan, Andrew},
  title     = {Option Tracing: Beyond Correctness Analysis in Knowledge Tracing},
  booktitle = {Proceedings of the 22nd International Conference on Artificial Intelligence in Education},
  pages     = {137--149},
  year      = {2021},
  doi       = {10.1007/978-3-030-78292-4_12},
  note      = {\url{https://doi.org/10.1007/978-3-030-78292-4_12}},
}

@inproceedings{vie_kashima_2023,
  author    = {Vie, Jill-J{\^e}nn and Kashima, Hisashi},
  title     = {Deep Knowledge Tracing Is an Implicit Dynamic Multidimensional Item Response Theory Model},
  booktitle = {Proceedings of the 31st International Conference on Computers in Education},
  year      = {2023},
  note      = {\url{https://arxiv.org/abs/2309.12334}},
}

@inproceedings{shen_dimkt_2022,
  author    = {Shen, Shuanghong and Huang, Zhenya and Liu, Qi and Su, Yu and Wang, Shijin and Chen, Enhong},
  title     = {Assessing Student's Dynamic Knowledge State by Exploring the Question Difficulty Effect},
  booktitle = {Proceedings of the 45th International ACM SIGIR Conference on Research and Development in Information Retrieval},
  pages     = {427--437},
  year      = {2022},
  doi       = {10.1145/3477495.3531939},
  note      = {\url{https://doi.org/10.1145/3477495.3531939}},
}

@inproceedings{yin_dtransformer_2023,
  author    = {Yin, Yu and Dai, Le and Huang, Zhenya and Shen, Shuanghong and Wang, Fei and Liu, Qi and Chen, Enhong and Li, Xin},
  title     = {Tracing Knowledge Instead of Patterns: Stable Knowledge Tracing with Diagnostic Transformer},
  booktitle = {Proceedings of the ACM Web Conference 2023},
  pages     = {855--864},
  year      = {2023},
  doi       = {10.1145/3543507.3583255},
  note      = {\url{https://doi.org/10.1145/3543507.3583255}},
}

@inproceedings{tsutsumi_independent_2021,
  author    = {Tsutsumi, Emiko and Kinoshita, Ryo and Ueno, Maomi},
  title     = {{Deep-IRT} with Independent Student and Item Networks},
  booktitle = {Proceedings of the 14th International Conference on Educational Data Mining},
  pages     = {510--517},
  year      = {2021},
  note      = {\url{https://educationaldatamining.org/EDM2021/virtual/static/pdf/EDM21_paper_43.pdf}},
}

@article{minn_ikt_2022,
  author    = {Minn, Sein and Vie, Jill-Jenn and Takeuchi, Koh and Kashima, Hisashi and Zhu, Feida},
  title     = {Interpretable Knowledge Tracing: Simple and Efficient Student Modeling with Causal Relations},
  journal   = {Proceedings of the AAAI Conference on Artificial Intelligence},
  volume    = {36},
  number    = {11},
  pages     = {12810--12818},
  year      = {2022},
  doi       = {10.1609/aaai.v36i11.21560},
  note      = {\url{https://doi.org/10.1609/aaai.v36i11.21560}},
}

@inproceedings{zhou_psikt_2024,
  author    = {Zhou, Hanqi and Bamler, Robert and Wu, Charley M. and Tejero-Cantero, {\'A}lvaro},
  title     = {Predictive, Scalable and Interpretable Knowledge Tracing on Structured Domains},
  booktitle = {International Conference on Learning Representations},
  year      = {2024},
  url       = {https://openreview.net/forum?id=NgaLU2fP5D},
  note      = {\url{https://openreview.net/forum?id=NgaLU2fP5D}},
}

@inproceedings{yang_gikt_2020,
  author    = {Yang, Yang and Shen, Jian and Qu, Yanru and Liu, Yunfei and Wang, Kerong and Zhu, Yaoming and Zhang, Weinan and Yu, Yong},
  title     = {{GIKT}: A Graph-Based Interaction Model for Knowledge Tracing},
  booktitle = {Machine Learning and Knowledge Discovery in Databases (ECML PKDD 2020)},
  series    = {Lecture Notes in Computer Science},
  volume    = {12457},
  pages     = {299--315},
  year      = {2021},
  doi       = {10.1007/978-3-030-67658-2_18},
  note      = {\url{https://doi.org/10.1007/978-3-030-67658-2_18}},
}

@article{chen_qikt_2023,
  author    = {Chen, Jiahao and Liu, Zitao and Huang, Shuyan and Liu, Qiongqiong and Luo, Weiqi},
  title     = {Improving Interpretability of Deep Sequential Knowledge Tracing Models with Question-Centric Cognitive Representations},
  journal   = {Proceedings of the AAAI Conference on Artificial Intelligence},
  volume    = {37},
  number    = {12},
  pages     = {14196--14204},
  year      = {2023},
  doi       = {10.1609/aaai.v37i12.26661},
  note      = {\url{https://doi.org/10.1609/aaai.v37i12.26661}},
}

\clearpage
\appendix
\section{Supporting Results}\label{app:supporting}

\subsection{Additional Item-Order Comparisons}\label{app:order-comparison}

\begin{figure}[!htbp]
\centering
\includegraphics[width=\textwidth]{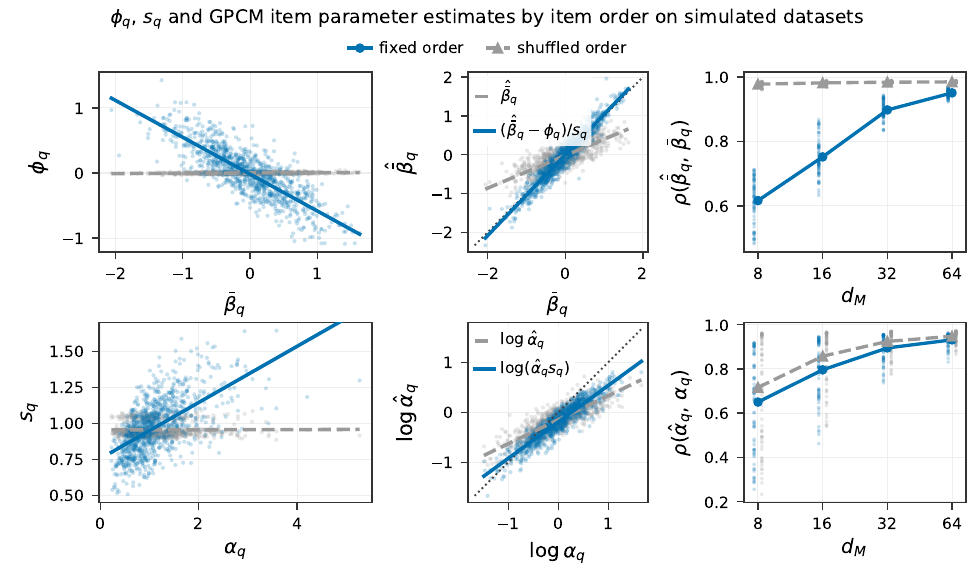}
\Description{Six GPCM panels compare ability offsets, gains, corrected item
estimates, and recovery across parameter key widths. Fixed-order offsets are
negatively associated with mean thresholds, and gains are positively
associated with discrimination. Corrected estimates are closer to the identity
line. Shuffled-order mean threshold recovery stays high across widths, while
discrimination recovery increases.}
\caption{Item order and GPCM parameter estimation. The left column plots
\(\phi_q\) and \(s_q\) against the mean step threshold \(\bar\beta_q\) and
discrimination \(\alpha_q\). The middle column compares the uncorrected and
affine-corrected estimates. The right column varies \(\dM\) while \(\dS=8\).}
\label{fig:channels-gpcm}
\end{figure}

\begin{figure}[!htbp]
\centering
\includegraphics[width=\textwidth]{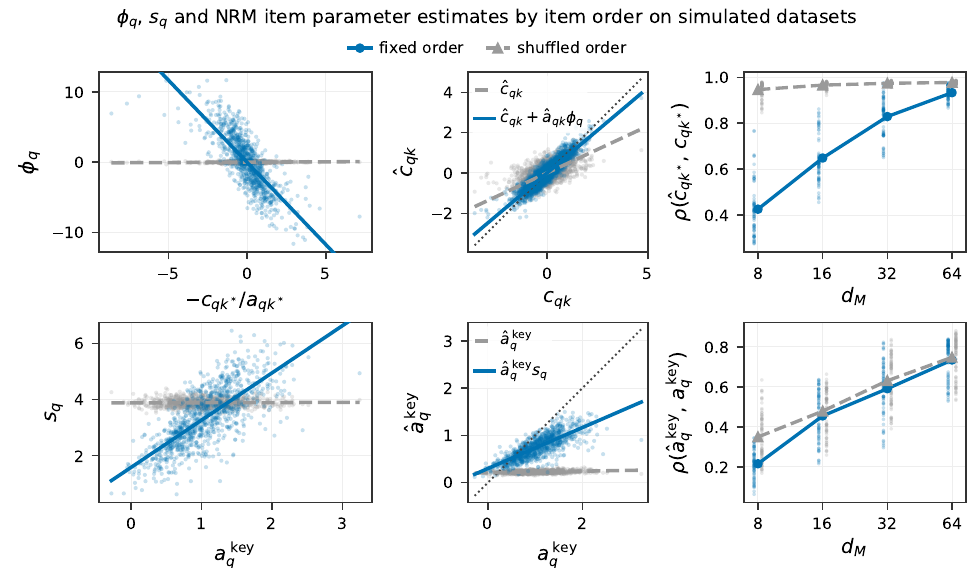}
\Description{Six NRM panels show a negative fixed-order association between
ability offset and correct-option location, and a positive association between
gain and keyed option slope contrast. Affine correction improves the relation
between estimated and generating intercepts and contrasts without exact
recovery. Under shuffled order, intercept recovery remains high across widths
while keyed contrast recovery increases.}
\caption{Item order and NRM parameter estimation. The left column uses the
correct option location \(-c_{q,k_q^\star}/a_{q,k_q^\star}\) and keyed option
slope contrast \(a_q^{\mathrm{key}}\). The middle column applies
Eq.~\eqref{eq:nrm_correction}. The right column uses
\(a_q^{\mathrm{key}}\). The location panels exclude 35 items with generating
\(a_{q,k_q^\star}\leq0.2\) to avoid dividing by a correct option slope near
zero.}
\label{fig:channels-nrm}
\end{figure}

\begin{figure}[!htbp]
\centering
\includegraphics[width=\textwidth]{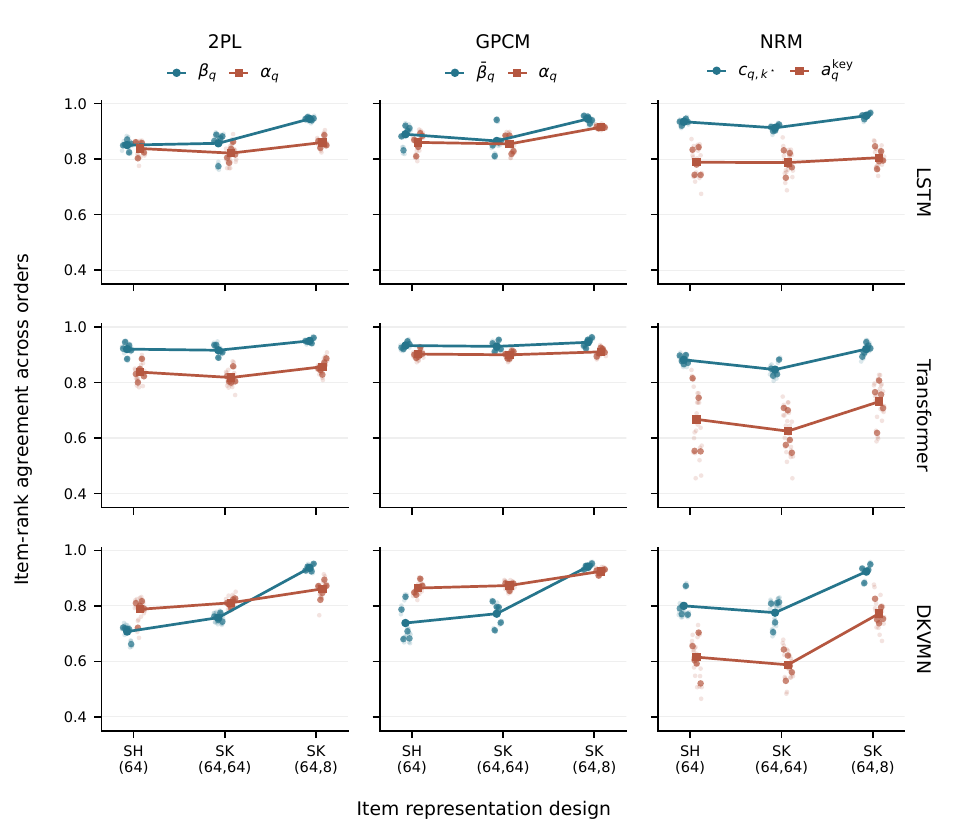}
\Description{Nine panels compare item rankings from fixed and shuffled orders
for SH(64), SK(64,64), and SK(64,8). Blue curves show difficulty, mean threshold,
or correct-option intercept, and orange curves show discrimination or keyed
contrast. SK(64,8) has higher mean cross-order agreement than SH(64) for both
parameter groups in every panel. SK(64,64) does not consistently improve agreement.}
\caption{Spearman agreement \(\rho_\xi^{\mathrm{ord}}\) between item estimates
from fixed and shuffled orders. Colors identify the parameters. Lines join the
means for \(\mathrm{SH}(64)\), \(\mathrm{SK}(64,64)\), and
\(\mathrm{SK}(64,8)\). Faint points are paired runs and darker points average
the five folds within each generated dataset.}
\label{fig:order-dependence}
\end{figure}

Figures~\ref{fig:channels-gpcm} and~\ref{fig:channels-nrm} present the six-panel
item-order and affine-correction comparisons for the GPCM and NRM corresponding
to the 2PL results in Figure~\ref{fig:channels-2pl} and Table~\ref{tab:traj}.
Figure~\ref{fig:order-dependence} then compares the estimated rankings of the
same 200 items across fixed and shuffled administrations
(\(\rho_\xi^{\mathrm{ord}}\)) for \(\mathrm{SH}(64)\), \(\mathrm{SK}(64,64)\),
and \(\mathrm{SK}(64,8)\) across all 25 matched dataset and fold pairs per
design. Larger \(\rho_\xi^{\mathrm{ord}}\) indicates that the learned item
rankings change less across administrations.

\subsection{Independent Affine Checks}\label{app:affine-checks}

We check the affine approximation separately for each trained model at
SK(8,8) and SK(64,8), under both item orders, giving 900 runs. Within each run,
we randomly divide the 400 test learners into two halves, fit on one half and
evaluate on the other, then exchange the halves. We compare a common affine
fit, a joint fit of gain and intercept for each item, and the gain combined
with the demeaned offset. Corrections are computed within each run before
averaging. Generating abilities are used only for this comparison and no
model weights are updated.

Ability error is RMSE divided by the standard deviation of the test-half
ability estimates. Logit error subtracts each response's mean over categories
before computing RMSE. Lower values indicate a closer approximation.
Item-specific fits reduce ability error under fixed order but not shuffled
order. Their remaining error means that the fitted transformations are not
exact response-equivalent models.

Table~\ref{tab:affine-generalization} compares the three affine fits on
learners not used to estimate their coefficients. All 200 items satisfy the
minimum exposure count in every fitting half. Nine item fits across eight
of the 1,800 fitting halves have nonpositive gains, all under fixed-order
NRM. These items remain in the prediction-error calculation and are excluded
from both recovery values in Table~\ref{tab:affine-recovery}.

\begin{table}[!htbp]
\centering
\small
\caption{Affine fit error on test learners not used to fit the transformation. The three middle columns report ability RMSE normalized by the test-half standard deviation, and Logit RMSE uses the itemwise fit on category-centered logits. Each row averages 150 runs, with the two fitting directions averaged within run before folds and datasets.}
\label{tab:affine-generalization}
\begin{tabular}{llrrrr}
\toprule
Model & Order & Common & Itemwise & Demeaned & Logit RMSE \\
\midrule
2PL & Fixed & 0.666 & 0.444 & 0.492 & 0.248 \\
2PL & Shuffled & 0.554 & 0.564 & 0.578 & 0.263 \\
GPCM & Fixed & 0.470 & 0.334 & 0.360 & 0.395 \\
GPCM & Shuffled & 0.371 & 0.378 & 0.384 & 0.408 \\
NRM & Fixed & 0.721 & 0.477 & 0.533 & 0.276 \\
NRM & Shuffled & 0.618 & 0.628 & 0.637 & 0.291 \\
\bottomrule
\end{tabular}
\end{table}

\begin{table}[!htbp]
\centering
\small
\caption{Run-level recovery before and after the itemwise affine correction. Here $\zeta_q$ is 2PL difficulty, GPCM mean step threshold, or NRM correct-option intercept. The parameter $\xi_q$ is discrimination $\alpha_q$ for 2PL and GPCM or the keyed option slope contrast $a_q^{\mathrm{key}}$ for NRM. Items with nonpositive fitted gains are excluded from the recovery correlations but retained in the prediction-error check.}
\label{tab:affine-recovery}
\begin{tabular}{llrrrr}
\toprule
Model & Order & $\rho_{\zeta}$ & $\rho_{\zeta}^{\rm corr}$ & $\rho_{\xi}$ & $\rho_{\xi}^{\rm corr}$ \\
\midrule
2PL & Fixed & 0.790 & 0.977 & 0.732 & 0.784 \\
2PL & Shuffled & 0.974 & 0.970 & 0.743 & 0.720 \\
GPCM & Fixed & 0.784 & 0.980 & 0.791 & 0.883 \\
GPCM & Shuffled & 0.982 & 0.978 & 0.832 & 0.820 \\
NRM & Fixed & 0.679 & 0.974 & 0.475 & 0.726 \\
NRM & Shuffled & 0.962 & 0.959 & 0.548 & 0.519 \\
\bottomrule
\end{tabular}
\end{table}

\subsection{DKVMN Ability Readout}\label{app:dkvmn-readout}

\begin{figure}[!htbp]
\centering
\includegraphics[width=\textwidth]{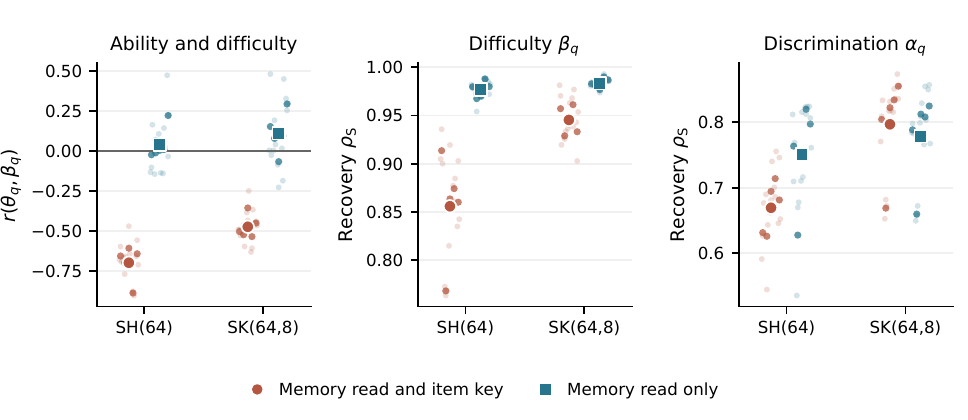}
\Description{Three panels compare DKVMN with a memory-and-item readout and a
memory-only readout under SH(64) and SK(64,8). Removing the item key moves the
ability-difficulty correlation close to zero and improves difficulty recovery
for both designs. Discrimination recovery increases for SH but decreases
slightly for SK. Faint points retain the individual runs.}
\caption{DKVMN ability readout and parameter recovery under random item
order. The first panel reports Pearson correlation between the per-item
mean ability estimate \(\overline{\hat\theta}_q\) and generating difficulty.
The remaining panels report Spearman recovery. Faint points show 15 runs per
condition, darker points average the three initializations within each of
five datasets, and large markers show their mean.}
\label{fig:dkvmn-summary}
\end{figure}

The supplementary study uses five generated 2PL datasets with 800 learners,
60 items, 60 responses per learner, and fixed generating abilities. Item order
is balanced and random. Each dataset has three initialization seeds and a
fixed 80/20 learner split. Training uses 120 full-batch Adam epochs at
\(10^{-2}\). The memory has 20 slots, value width 32, and normally initialized
values with standard deviation 0.1. Both conditions read memory before
updating it, using the current item for addressing. The ability readout
receives the memory read with or without the concatenated item key. This
differs from the history-only ability readout in the width study and scores
all 60 responses of each test learner rather than the final ten.
Figure~\ref{fig:dkvmn-summary} compares the resulting ability associations
and item parameter recovery.

With the item key present, the correlation between ability and difficulty is
\(-0.698\) for SH and \(-0.474\) for SK. Removing it changes these values to
0.040 and 0.110. Difficulty recovery increases from 0.856 to 0.977 for SH and
from 0.945 to 0.983 for SK. Discrimination recovery increases from 0.669 to
0.751 for SH but decreases from 0.797 to 0.779 for SK. The smaller gap between
SH and SK therefore accompanies a weaker association between ability and difficulty,
consistent with an effect of the ability input on item parameter estimation.

Mean test log loss increases by 0.0014 in SH and 0.0010 in SK, while accuracy
falls by 0.0024 and 0.0014. Parameter counts fall from 17,651 to 15,603 in SH
and from 9,339 to 9,083 in SK. The comparison changes both direct item input
and readout size. It does not remove item-dependent memory addressing or
establish that removing the key improves every outcome.

\subsection{Probability Quality and Item Ranks}\label{app:probability-checks}

Tables~\ref{tab:sim-prediction-loss} and~\ref{tab:real-rank-prediction}
accompany the recovery and offline-agreement comparisons with test log loss.
Each difference uses matched models on the same scored responses.
Table~\ref{tab:real-rank-prediction} also reports direct agreement between the
SH and SK item rankings, without using the offline reference.

Saved trajectories reconstruct the 2PL and NRM probabilities. The 55 TIMSS
SH(64) replays reproduce the original item parameters and accuracy exactly.
The simulated GPCM values use five supplementary runs per model, one fold
from each generating dataset, because the original exports sorted the
thresholds. These supplementary runs retain raw thresholds and use the stated
120-epoch training settings. Their results are not substituted for the
original recovery cells.

Mean log loss decreases in eight of nine simulated comparisons. On real data,
the LSTM and Transformer change in opposite directions on EdNet 2PL. SK lowers
LSTM log loss by 0.076 but raises Transformer log loss by 0.068. Matched test
accuracy therefore does not imply equivalent probability predictions across designs.

\begin{figure}[!htbp]
\centering
\includegraphics[width=\textwidth]{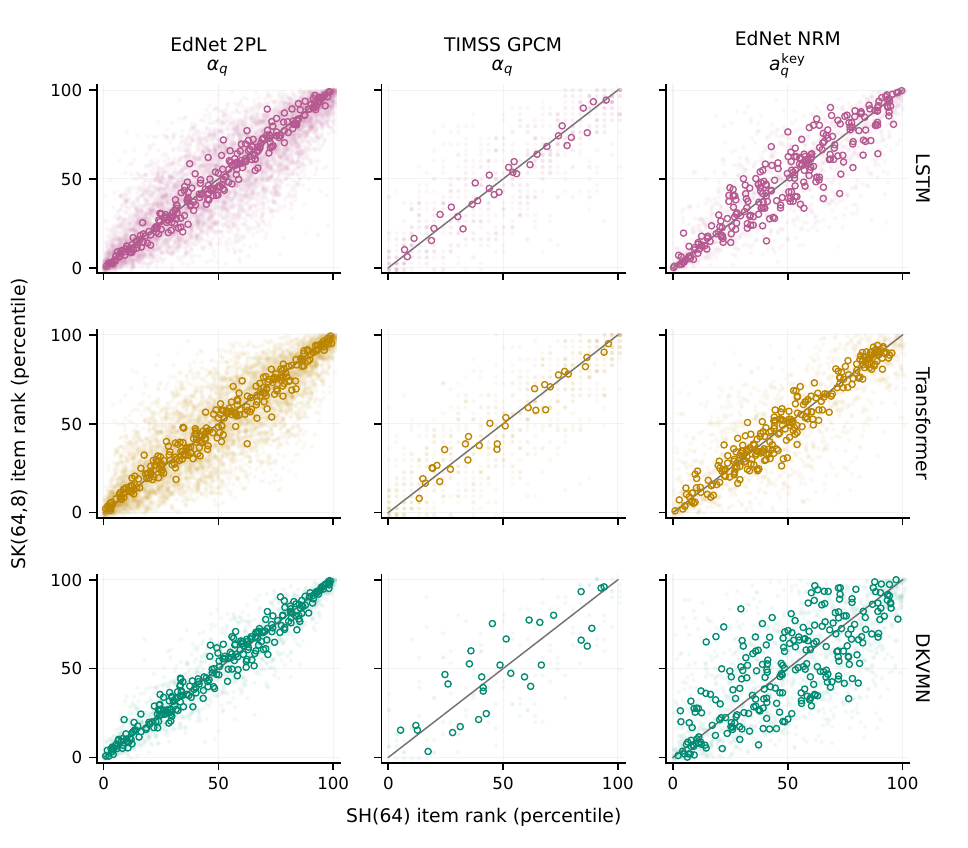}
\Description{Nine scatterplots place each item's SK(64,8) percentile rank
against its SH(64) rank. Points on the diagonal have equal ranks. EdNet 2PL
and TIMSS GPCM show close average rankings with run-level spread, while EdNet
NRM has larger changes, especially for DKVMN. Open circles show average item
ranks and faint points show paired runs.}
\caption{Ranks assigned to the same items by SH(64) and SK(64,8). Faint
points are paired runs and open circles average each item's ranks over pairs.
The diagonal denotes equal ranks. LSTM and Transformer have 25 pairs for
2PL and GPCM and five for NRM. DKVMN has five pairs in each column.}
\label{fig:item-ranks}
\end{figure}

For the direct real-item comparison, ranks are scaled from 0 to 100 within
each model and run. The reported rank change is the mean across runs of the
median absolute item-rank difference. All 31 TIMSS or 250 EdNet items enter,
without filtering by the offline reference. These differences measure
dependence on the representation design, not which model is correct.
Figure~\ref{fig:item-ranks} plots the corresponding item ranks.

\begin{table}[!htbp]
\centering
\small
\caption{Test prediction changes from SH(64) to SK(64,8). Means weight the five generated datasets equally. The last two columns report the minimum and maximum dataset-mean log-loss changes across the five generated datasets. The 2PL and NRM reuse 25 runs per model. GPCM uses five supplementary runs per model, one fold from each dataset, with raw unsorted thresholds retained.}
\label{tab:sim-prediction-loss}
\begin{tabular}{llrrrrr}
\toprule
Model & Encoder & $n$ & $\Delta\mathrm{Acc}$ & $\Delta\mathrm{NLL}$ & Minimum & Maximum \\
\midrule
2PL & LSTM & 25 & +0.0007 & -0.0022 & -0.0030 & -0.0007 \\
2PL & Transformer & 25 & +0.0004 & +0.0007 & -0.0016 & +0.0034 \\
2PL & DKVMN & 25 & +0.0022 & -0.0033 & -0.0042 & -0.0021 \\
GPCM & LSTM & 5 & +0.0138 & -0.0307 & -0.0374 & -0.0261 \\
GPCM & Transformer & 5 & +0.0040 & -0.0060 & -0.0178 & +0.0031 \\
GPCM & DKVMN & 5 & +0.0053 & -0.0086 & -0.0125 & -0.0060 \\
NRM & LSTM & 25 & +0.0024 & -0.0048 & -0.0073 & -0.0027 \\
NRM & Transformer & 25 & +0.0033 & -0.0059 & -0.0089 & -0.0003 \\
NRM & DKVMN & 25 & +0.0052 & -0.0097 & -0.0110 & -0.0084 \\
\bottomrule
\end{tabular}
\end{table}

\begin{table}[!htbp]
\centering
\small
\caption{Matched SK(64,8) minus SH(64) comparisons. Rank change is the mean across runs of the median absolute item-rank change, in percentile points. The ranks use all 31 TIMSS or 250 EdNet items. Here $n$ denotes the number of paired training runs, and positive $\Delta\mathrm{NLL}$ indicates higher test log loss.}
\label{tab:real-rank-prediction}
\begin{tabular}{llrrrrr}
\toprule
Study & Encoder & $n$ & $\rho_{\rm SH,SK}$ & Rank change & $\Delta\mathrm{Acc}$ & $\Delta\mathrm{NLL}$ \\
\midrule
EdNet 2PL & LSTM & 25 & 0.858 & 8.7 & +0.0211 & -0.0765 \\
EdNet 2PL & Transformer & 25 & 0.861 & 8.3 & -0.0179 & +0.0681 \\
EdNet 2PL & DKVMN & 5 & 0.936 & 5.6 & +0.0034 & -0.0055 \\
TIMSS GPCM & LSTM & 25 & 0.822 & 10.1 & -0.0016 & +0.0031 \\
TIMSS GPCM & Transformer & 25 & 0.824 & 10.5 & -0.0009 & +0.0014 \\
TIMSS GPCM & DKVMN & 5 & 0.724 & 13.3 & +0.0061 & -0.0070 \\
EdNet NRM & LSTM & 5 & 0.832 & 8.8 & +0.0105 & -0.0273 \\
EdNet NRM & Transformer & 5 & 0.789 & 11.2 & -0.0006 & +0.0005 \\
EdNet NRM & DKVMN & 5 & 0.670 & 14.1 & +0.0024 & -0.0099 \\
\bottomrule
\end{tabular}
\end{table}

\subsection{Item Representation Width Study}\label{app:width-study}

\begin{figure}[!htbp]
\centering
\includegraphics[width=\textwidth]{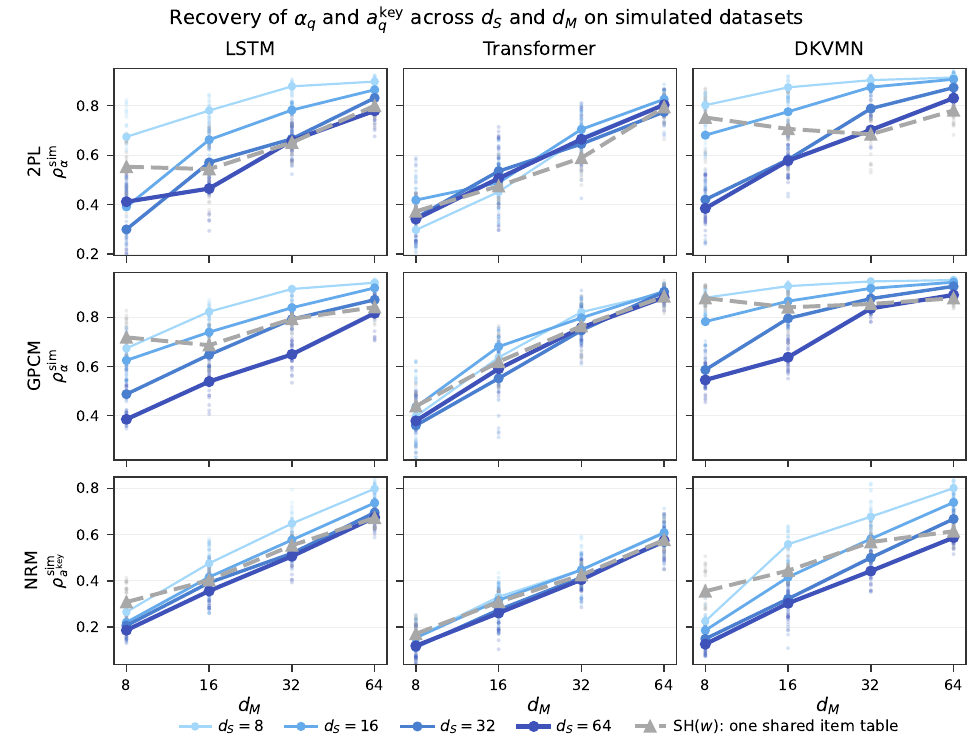}
\Description{Nine line plots report absolute simulated parameter recovery
across widths. Blue lines hold sequence width fixed, while the gray dashed
line varies the shared width. Recovery rises as parameter keys widen, and
thin sequence representations give higher recovery at the widest key.
The vertical levels differ across encoders and response models.}
\caption{Absolute recovery across \(\dM\) and \(\dS\) on simulated data.
Blue lines hold \(\dS\) fixed while \(\dM\) varies. The gray dashed line is
SH as \(w\) varies. The NRM row uses \(a_q^{\mathrm{key}}\).}
\label{fig:app-width-sim}
\end{figure}

\begin{figure}[!htbp]
\centering
\includegraphics[width=\textwidth]{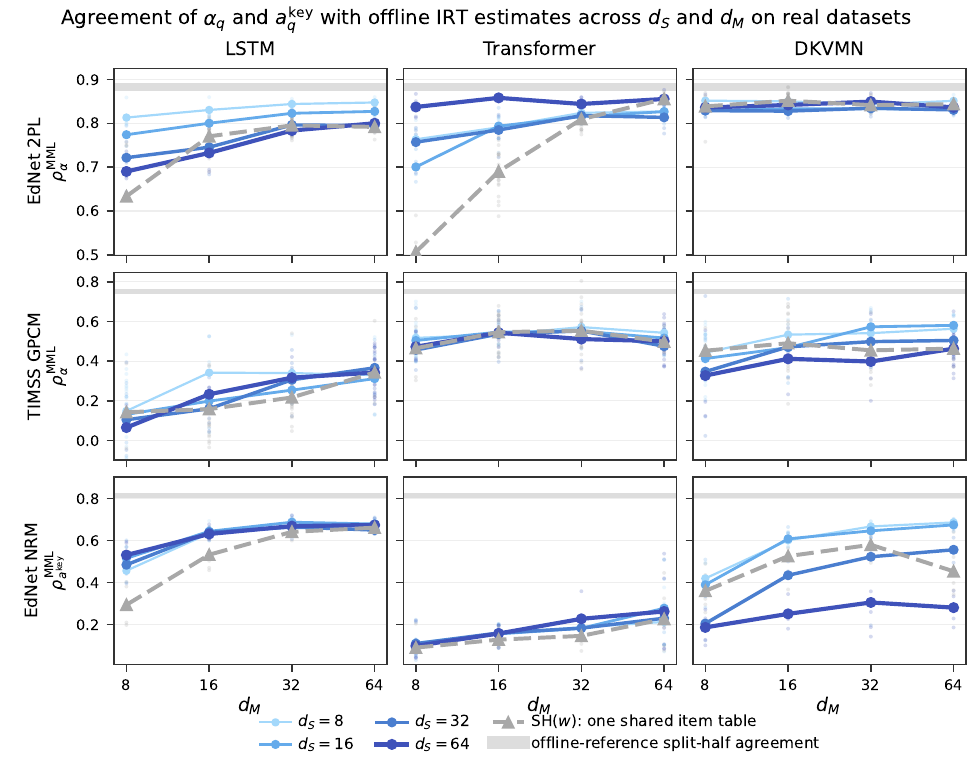}
\Description{Nine line plots report absolute agreement with offline item
parameter estimates. The DKVMN EdNet 2PL curves are nearly flat, while LSTM
agreement increases with parameter key width. TIMSS and EdNet NRM show
encoder-dependent width effects and run-level spread. Gray horizontal bands
mark split-half agreement between offline estimates.}
\caption{Absolute agreement with the offline IRT reference across \(\dM\)
and \(\dS\) on real data. Encodings follow
Figure~\ref{fig:app-width-sim}. The gray band gives the agreement between two
offline estimates from separate response halves.}
\label{fig:app-width-real}
\end{figure}

\begin{figure}[!htbp]
\centering
\includegraphics[width=\textwidth]{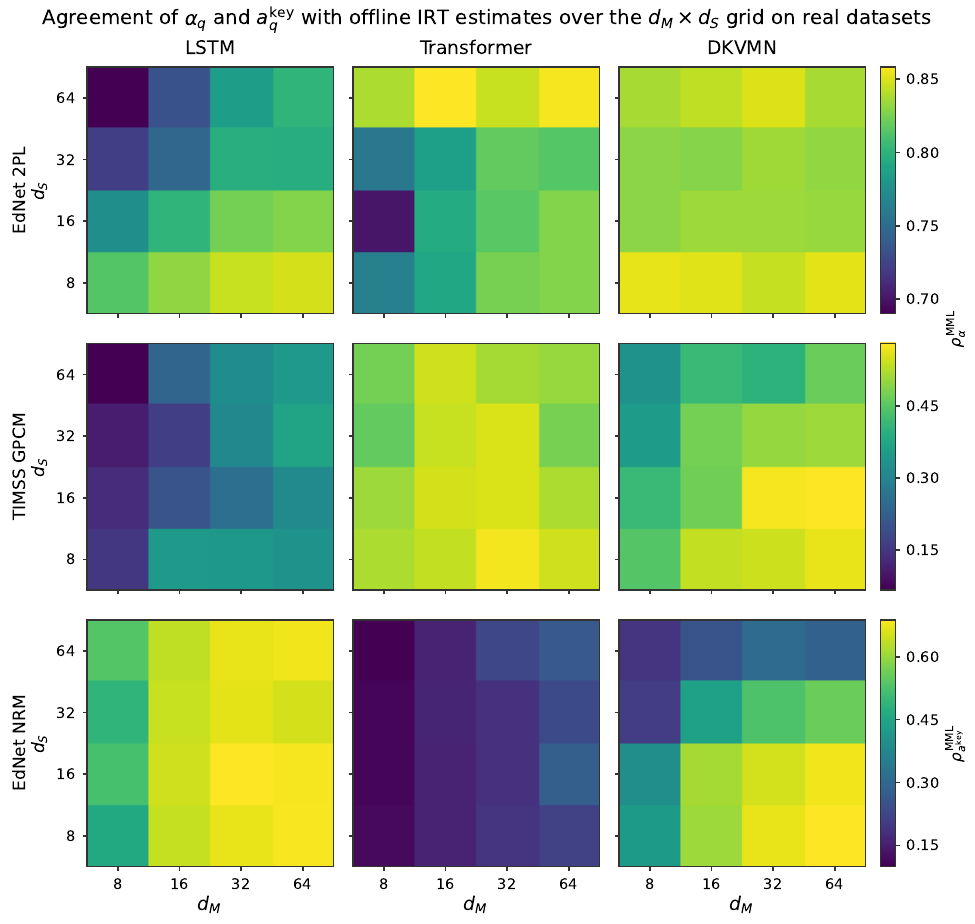}
\Description{Nine heatmaps show offline-reference agreement over the full
parameter-key and sequence-width grid. Brightness represents agreement, with
a common scale across encoders within each dataset and response model.
Larger keys often improve agreement, but the strongest sequence-width choices
differ across the nine panels.}
\caption{Agreement with the offline IRT reference over the complete SK width
grid on real data. Color limits are shared across encoders within a dataset
and response model.}
\label{fig:app-heatmap-real}
\end{figure}

Figures~\ref{fig:width-sim-rel} and~\ref{fig:width-real-rel} in
Section~\ref{sec:results_width} report changes relative to \(\mathrm{SH}(8)\),
\(\Delta_{\mathrm{SH}(8)}\rho_\xi\), and Figure~\ref{fig:app-heatmap-sim} shows
the complete simulated SK width grid. Figures~\ref{fig:app-width-sim}
and~\ref{fig:app-width-real} below retain the absolute recovery and offline
agreement curves, and Figure~\ref{fig:app-heatmap-real} shows the complete SK
width grid on the real-data benchmarks.

\FloatBarrier

\subsection{Agreement Across Training Runs}\label{app:run-agreement}

\begin{figure}[!htbp]
\centering
\includegraphics[width=\textwidth]{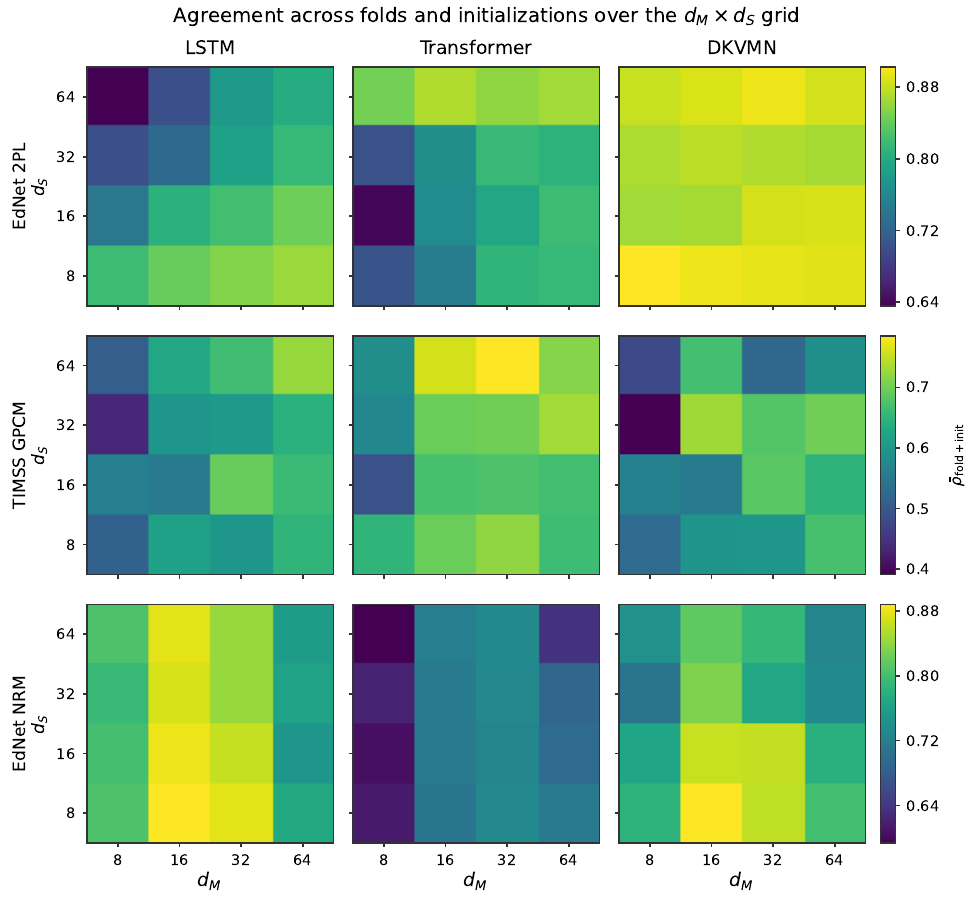}
\Description{Nine heatmaps show pairwise rank agreement between training runs
over parameter key and sequence widths. Brighter tiles indicate more similar
rankings. EdNet DKVMN and LSTM generally have higher agreement than the
Transformer NRM panels. TIMSS shows more irregular width effects. Both the
data split and initialization vary across compared runs.}
\caption{Agreement among item parameter estimates from training runs with
different folds and initializations. Each tile averages pairwise Spearman
correlations within one SK width combination. Color limits are shared across
encoders within a dataset and response model.}
\label{fig:app-heatmap-runs}
\end{figure}

Figure~\ref{fig:app-heatmap-runs} compares item parameter rankings across
training runs with different folds and initializations.

\subsection{Complete Matched Comparisons}
\label{app:complete-comparisons}

Tables~\ref{tab:sim-gaps} and~\ref{tab:real-gaps} retain the comparisons used
in Section~\ref{sec:results_width}. Tables~\ref{tab:sim-gaps-full}
and~\ref{tab:real-gaps-full} report the complete baseline comparisons and,
for simulation, the bootstrap intervals.

\FloatBarrier

\begin{sidewaystable}[p]
\centering
\caption{Complete simulated test accuracy and recovery comparisons for
SK(64,8). Entries average 25 training runs and report 95\% bootstrap intervals
clustered by generated dataset. Cost is the trainable parameter ratio
SK(64,8)/SH(64). Leading zeros are omitted.}
\label{tab:sim-gaps-full}
\scriptsize
\setlength{\tabcolsep}{1.6pt}
\resizebox{\linewidth}{!}{%
\begin{tabular}{@{}ll cccc cccc c@{}}
\toprule
 & & \multicolumn{4}{c}{Test accuracy}
 & \multicolumn{4}{c}{$\rho$, $\alpha_q$ or $a^{\,\mathrm{key}}_q$} & Cost \\
\cmidrule(lr){3-6} \cmidrule(lr){7-10} \cmidrule(l){11-11}
 & & \textbf{SK $(64,8)$} & vs SH(8) & vs SH(64) & vs SK $(8,64)$
 & \textbf{SK $(64,8)$} & vs SH(8) & vs SH(64) & vs SK $(8,64)$
 & Cost ratio \\
\midrule
2PL & LSTM & .714 [.709,.719] & +.003 [+.002,+.003] & +.001 [+.000,+.001] & +.004 [+.002,+.005] & .898 [.884,.911] & +.344 [+.279,+.416] & +.098 [+.071,+.125] & +.485 [+.414,+.587] & .62 \\
2PL & Transformer & .713 [.707,.717] & +.004 [+.003,+.006] & +.000 [-.001,+.002] & +.001 [-.002,+.003] & .806 [.785,.826] & +.433 [+.405,+.469] & +.011 [-.032,+.062] & +.464 [+.417,+.515] & .96 \\
2PL & DKVMN & .715 [.709,.721] & -.000 [-.002,+.002] & +.002 [+.002,+.003] & +.002 [+.001,+.003] & .914 [.901,.928] & +.162 [+.098,+.246] & +.133 [+.108,+.158] & +.529 [+.468,+.576] & .68 \\
\midrule
GPCM & LSTM & .502 [.499,.506] & +.015 [+.011,+.018] & +.008 [+.006,+.010] & +.028 [+.016,+.040] & .941 [.936,.946] & +.223 [+.176,+.269] & +.099 [+.071,+.128] & +.557 [+.451,+.693] & .62 \\
GPCM & Transformer & .492 [.489,.497] & +.033 [+.025,+.044] & +.001 [-.001,+.003] & +.028 [+.019,+.039] & .900 [.879,.920] & +.463 [+.444,+.478] & +.015 [-.001,+.029] & +.522 [+.418,+.658] & .96 \\
GPCM & DKVMN & .502 [.500,.506] & +.006 [+.003,+.009] & +.003 [+.001,+.005] & +.013 [+.010,+.016] & .952 [.947,.955] & +.073 [+.042,+.105] & +.074 [+.049,+.092] & +.407 [+.366,+.446] & .68 \\
\midrule
NRM & LSTM & .498 [.488,.511] & +.016 [+.011,+.020] & +.002 [+.000,+.005] & +.030 [+.026,+.033] & .798 [.787,.809] & +.490 [+.424,+.559] & +.126 [+.101,+.151] & +.611 [+.593,+.627] & .62 \\
NRM & Transformer & .489 [.481,.497] & +.040 [+.035,+.044] & +.003 [-.000,+.007] & +.043 [+.035,+.049] & .604 [.570,.637] & +.433 [+.402,+.463] & +.027 [+.003,+.052] & +.485 [+.441,+.532] & .96 \\
NRM & DKVMN & .499 [.489,.509] & +.005 [+.002,+.006] & +.005 [+.004,+.007] & +.017 [+.014,+.020] & .800 [.781,.819] & +.446 [+.371,+.520] & +.186 [+.163,+.204] & +.674 [+.665,+.687] & .68 \\
\bottomrule
\end{tabular}%
}
\end{sidewaystable}

\begin{sidewaystable}[p]
\centering
\scriptsize
\caption{Complete real-data comparisons for SK(64,8). Levels average all available runs. Each difference uses only matching seed and fold indices in the two models. Comparisons with SH(64) contain 25 pairs for LSTM and Transformer on 2PL and GPCM, and five otherwise. Comparisons with SH(8) and SK(8,64) contain five pairs. Cost is the parameter ratio to SH(64).}
\label{tab:real-gaps-full}
\begin{tabular}{llrrrrrrrrr}
\toprule
Study & Encoder & $\mathrm{Acc}_{\rm SK}$ & vs SH(8) & vs SH(64) & vs SK(8,64) & $\rho_{\rm SK}^{\rm MML}$ & vs SH(8) & vs SH(64) & vs SK(8,64) & Cost \\
\midrule
EdNet 2PL & LSTM & 0.641 & +0.000 & +0.021 & +0.028 & 0.846 & +0.219 & +0.043 & +0.162 & 0.46 \\
EdNet 2PL & Transformer & 0.617 & +0.002 & -0.018 & -0.022 & 0.827 & +0.326 & -0.031 & -0.009 & 0.77 \\
EdNet 2PL & DKVMN & 0.653 & -0.002 & +0.003 & +0.005 & 0.856 & +0.012 & +0.006 & +0.013 & 0.48 \\
TIMSS GPCM & LSTM & 0.579 & -0.002 & -0.002 & +0.000 & 0.325 & +0.177 & +0.027 & +0.263 & 0.35 \\
TIMSS GPCM & Transformer & 0.580 & +0.001 & -0.001 & +0.002 & 0.518 & +0.086 & +0.022 & +0.064 & 0.89 \\
TIMSS GPCM & DKVMN & 0.584 & +0.001 & +0.006 & +0.007 & 0.545 & +0.122 & +0.090 & +0.232 & 0.37 \\
EdNet NRM & LSTM & 0.590 & -0.002 & +0.010 & +0.004 & 0.681 & +0.387 & +0.020 & +0.150 & 1.07 \\
EdNet NRM & Transformer & 0.590 & +0.001 & -0.001 & -0.000 & 0.213 & +0.122 & -0.014 & +0.114 & 1.09 \\
EdNet NRM & DKVMN & 0.596 & -0.000 & +0.002 & +0.002 & 0.686 & +0.326 & +0.233 & +0.499 & 1.08 \\
\bottomrule
\end{tabular}
\end{sidewaystable}

\FloatBarrier

\subsection{Trainable Parameter Counts}\label{app:param-counts}

\begin{figure}[!htbp]
\centering
\Description{Six panels show absolute parameter recovery or offline agreement
against relative trainable parameter count. Color distinguishes encoders,
solid circles represent SK, and dashed triangles represent SH. Increasing
width generally moves simulated recovery rightward. Real-data agreement
trends differ across encoders, while the cost of width depends on both the
encoder and response model.}
\includegraphics[width=\textwidth]{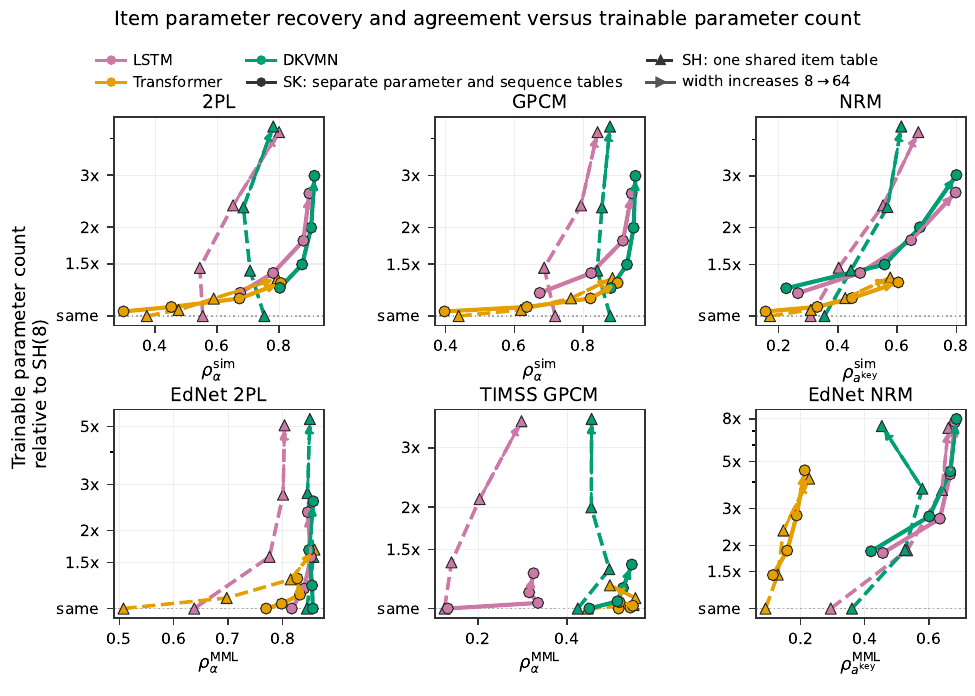}
\caption{Absolute item parameter recovery, offline agreement, and trainable parameter count.
The top row reports simulated recovery, and the bottom row reports agreement
with the offline IRT reference. Trainable parameter count is relative to
SH(8) on the logarithmic vertical axis. Series and arrow direction follow
Figure~\ref{fig:recovery-cost}.}
\label{fig:app-recovery-cost-absolute}
\end{figure}

Figure~\ref{fig:app-recovery-cost-absolute} reports absolute recovery and
offline agreement against model size. Table~\ref{tab:param-counts} gives
the real-data parameter counts.

\begin{table}[!htbp]
\centering
\caption{Trainable parameters, in thousands, for the real-data width study.
SK holds $\dS=8$ while $\dM$ varies. SH varies its shared width $w$. Counts
are exact and do not vary across training runs.}
\label{tab:param-counts}
\small
\setlength{\tabcolsep}{4pt}
\begin{tabular}{@{}llcccccccc@{}}
\toprule
 & & \multicolumn{4}{c}{SK, $\dS = 8$}
 & \multicolumn{4}{c}{SH} \\
\cmidrule(lr){3-6}\cmidrule(lr){7-10}
Dataset and model & Encoder & $\dM{=}8$ & 16 & 32 & 64
                & $w{=}8$ & 16 & 32 & 64 \\
\midrule
EdNet 2PL    & LSTM        & 10.5 & 12.5 & 16.5 & 24.6 & 10.5 & 16.5 & 28.7 & 53.0 \\
             & Transformer & 46.4 & 48.4 & 52.5 & 60.5 & 46.4 & 50.9 & 60.0 & 78.2 \\
             & DKVMN       & 8.9 & 10.9 & 15.0 & 23.0 & 8.9 & 14.1 & 24.8 & 47.7 \\
\addlinespace[2pt]
TIMSS GPCM   & LSTM        & 7.0 & 7.3 & 7.8 & 8.9 & 7.0 & 9.6 & 14.8 & 25.1 \\
             & Transformer & 42.9 & 43.2 & 43.7 & 44.8 & 42.9 & 44.0 & 46.1 & 50.3 \\
             & DKVMN       & 5.4 & 5.7 & 6.2 & 7.3 & 5.4 & 7.1 & 10.8 & 19.8 \\
\addlinespace[2pt]
EdNet NRM    & LSTM        & 74.1 & 107.9 & 175.5 & 310.8 & 40.3 & 76.2 & 148.0 & 291.6 \\
             & Transformer & 110.0 & 143.8 & 211.5 & 346.8 & 76.2 & 110.6 & 179.3 & 316.8 \\
             & DKVMN       & 72.5 & 106.3 & 174.0 & 309.3 & 38.7 & 73.7 & 144.1 & 286.3 \\
\bottomrule
\end{tabular}
\end{table}

\end{document}